\documentclass[lettersize,journal]{IEEEtran}
\usepackage{amsmath,amsfonts}
\usepackage{algorithmicx}
\usepackage{algpseudocode}
\usepackage{algorithm}
\usepackage{array}
\usepackage[caption=false,font=normalsize,labelfont=sf,textfont=sf]{subfig}
\usepackage{textcomp}
\usepackage{stfloats}
\usepackage{url}
\usepackage{verbatim}
\usepackage{graphicx}
\usepackage{cite}
\usepackage{acronym}
\usepackage{orcidlink}
\usepackage{threeparttable}
\usepackage{multirow}
\usepackage[normalem]{ulem}
\usepackage{xcolor} 

\newacro{HAR}{Human Activity Recognition}
\newacro{RadHAR}{Radar-based Human Activity Recognition}

\newacro{RF}{Radio Frequency}

\newacro{CW}{Continuous Wave}
\newacro{FMCW}{Frequency Modulated Continuous Wave}
\newacro{UWB}{Ultra-Wide Band}
\newacro{PRI}{Pulse Repetition Interval}

\newacro{STFT}{Short-Time Fourier Transform}
\newacro{SVD}{Singular Value Decomposition}
\newacro{ZOH}{Zero-Order Hold}

\newacro{DL}{Deep Learning}
\newacro{AdamW}{Adaptive Moment Estimation with decoupled Weight decay}

\newacro{1D}{One-Dimensional}
\newacro{2D}{Two-Dimensional}

\newacro{kNN}{k-Nearest Neighbors}
\newacro{SVM}{Support Vector Machine}

\newacro{CNN}{Convolutional Neural Network}
\newacro{RNN}{Recurrent Neural Network}
\newacro{Bi}{Bidirectional}
\newacro{LSTM}{Long Short-Term Memory}
\newacro{GRU}{Gated Recurrent Unit}

\newacro{Bi-LSTM}{Bidirectional Long Short-Term Memory}
\newacro{CNN-LSTM}{}

\newacro{ViT}{Vision Transformer}
\newacro{SSM}{State Space Model}
\newacro{ViM}{Vision Mamba}

\newacro{SiLU}{Sigmoid Linear Unit}
\newacro{FML}{Frequency Modulated Lightweight}
\newacro{LW}{Lightweight}

\newacro{DIAT}{Defence Institute of Advanced categorization problems and classification
Technology, India}
\newacro{CI4R}{Computational Intelligence for Radar}
\newacro{UoG20}{University of Glasgow in 2020}
\newacro{CP-Mamba}{Convolutional Projection-based Mamba}

\usepackage{lipsum}
\usepackage{fancyhdr}
\fancypagestyle{sec}{\lhead{\tmpx}}

\fancypagestyle{arXiv}
{
   \fancyhf{}
   \chead{\textcolor{gray}{This article has been accepted to the IEEE Transactions on Radar Systems (T-RS).}}
   \cfoot{ \fontsize{=9pt}{10pt}\selectfont  \textcolor{gray}{© 2026 IEEE. Personal use of this material is permitted. Permission from IEEE must be obtained for all other uses, in any current or future media, including reprinting/republishing this material for advertising or promotional purposes, creating new collective works, for resale or redistribution to servers or lists, or reuse of any copyrighted component of this work in other works}\\\fontsize{=10pt}{12pt}\selectfont\thepage}
   
}

\begin{document}

\title{RadMamba: Efficient Human Activity Recognition through a Radar-based Micro-Doppler-Oriented Mamba State-Space Model}

\author{Yizhuo~Wu\orcidlink{0009-0009-5087-7349}, ~\IEEEmembership{Student Member, IEEE},
        Francesco Fioranelli\orcidlink{0000-0001-8254-8093}, ~\IEEEmembership{Senior Member, IEEE} \\
    Chang~Gao*\orcidlink{0000-0002-3284-4078}, ~\IEEEmembership{Member, IEEE}
    \thanks{Yizhuo Wu, Francesco Fioranelli, and Chang Gao are with the Department of Microelectronics, Delft University of Technology, The Netherlands.}
    \thanks{*Corresponding author: Chang Gao (chang.gao@tudelft.nl)} \\}


% % column for its text to clear the IEEEpubid mark.

\maketitle

\begin{abstract}
% Radar-based \ac{HAR} is an attractive alternative to wearables and cameras because it preserves privacy and is robust to illumination and occlusions. However, dominant \ac{CNN}- and \ac{RNN}-based solutions are computation-heavy at deployment, and recent lightweight \ac{ViT} and \ac{SSM} variants still incur substantial complexity. We present \textit{RadMamba}, a parameter-efficient, micro-Doppler-oriented Mamba \ac{SSM} tailored to radar \ac{HAR}. RadMamba combines (i) channel fusion with downsampling, (ii) Doppler-aligned segmentation, and (iii) convolutional token projections to retain temporal–Doppler structure while reducing \#FLOP/Inf. Across three datasets, RadMamba matches the prior best 99.8\% accuracy on \texttt{\ac{DIAT}} with only 1/400 of the parameters, equals leading models' 92.0\% on \texttt{\ac{CI4R}} with about 1/10 of their parameters, and surpasses methods with far more parameters on \texttt{UoG2020} by at least 3\%, using only 6.7\,k parameters. Code: \url{https://github.com/lab-emi/AIRHAR}.

Radar-based \ac{HAR} is an attractive alternative to wearables and cameras because it preserves privacy, and is contactless and robust to occlusions. However, dominant \ac{CNN}- and \ac{RNN}-based solutions are computationally intensive at deployment, and recent lightweight \ac{ViT} and \ac{SSM} variants still exhibit substantial complexity. In this paper, we present \textit{RadMamba}, a parameter-efficient, micro-Doppler–oriented Mamba \ac{SSM} tailored to radar \ac{HAR} under on-sensor compute, latency, and energy constraints typical of distributed radar systems. RadMamba combines (i) channel fusion with downsampling, (ii) Doppler-aligned segmentation that preserves the physical continuity of Doppler over time, and (iii) convolutional token projections that better capture Doppler-span variations, thereby retaining temporal–Doppler structure while reducing the number of Floating-point Operations/Inference ($\#$FLOP/Inf.). Evaluated across three datasets with different radars and types of activities, RadMamba matches the prior best $99.8\%$ accuracy of a recent SSM-based model on the \ac{CW} radar dataset, while requiring only $1/400$ of its parameters. On a dataset of non-continuous activities with \ac{FMCW} radar, RadMamba remains competitive with leading $92.0\%$ results using about $1/10$ of the parameters, and on a continuous \ac{FMCW} radar dataset it surpasses methods with far more parameters by at least $3\%$, using only $6.7\,\mathrm{k}$ parameters. {\color{red}Code: \url{https://github.com/lab-emi/AIRHAR}}.

\end{abstract}


\section{Introduction}
\thispagestyle{arXiv}
\IEEEPARstart{H}{uman} Activity Recognition (HAR) technologies are central to healthcare, elderly care, smart homes, and security. Traditional HAR solutions rely on wearable sensors or cameras. While wearables can offer precise monitoring, they suffer from issues related to user compliance, battery life, and comfort. Camera-based approaches raise privacy concerns and are at high risk of degradation under challenging environmental conditions. Consequently, \ac{RadHAR} has emerged as a compelling alternative, leveraging active electromagnetic sensing to infer behaviors while preserving privacy, maintaining robustness to illumination and occlusions, and enabling through-wall sensing~\cite{survey2019,Survey-Continuous}.

Early \ac{RadHAR} studies progressed from simple, isolated activities~\cite{Kim2009,Mattew2016} to more realistic settings involving continuous activity monitoring~\cite{Erol2018,Li2019}, multi-modal and distributed sensing~\cite{multimodal2020,DatasetD}, and advanced classifiers. Conventional machine learning with hand-crafted micro-Doppler features (e.g., centroid, bandwidth, statistical descriptors, and \ac{SVD} components)~\cite{Mattew2016,Erol2018,Li2019,Ding2019} demonstrated proof of concept but struggled with scalability and generalization in continuous scenarios.

Deep learning improved accuracy by learning spatio-temporal representations directly from radar signatures. \ac{CNN}-based methods~\cite{Kim2016,CNN_TGRS,ComplexNN_TGRS,Yu2022}, \ac{RNN}-based methods~\cite{Bi-RNN,multimodal2020}, and hybrid \ac{CNN}-\ac{RNN} models~\cite{CNN-BiLSTM,CNN-LSTM,CNN-RNN-distributed} substantially advanced performance. However, they often have high computational cost and latency, which limits on-sensor deployment in resource-constrained radar nodes. More recently, transformer and \ac{SSM}-based architectures~\cite{Vim,FML-VIT,VitAIHAR2022,dosovitskiy2021an,ActivityMamba,Ren2025TPAMI} have shown strong modeling capacity and explored lightweight designs, but balancing parameter count with computational efficiency, measured by the number of floating-point operations per inference (\#FLOP/Inf.), remains challenging for mobile and distributed applications such as gesture interfaces~\cite{linardakis2025} and smart-home radar networks~\cite{DatasetD}.

Micro-Doppler signatures encode the evolution of Doppler frequency over time. Preserving this physical continuity is crucial for maintaining the high accuracy of sequential models; however, common vision-style rectangular patching and tokenization can disrupt Doppler–time ordering and waste computational resources on sparse, low-information regions. Moreover, edge radar deployments impose strict limits on memory, latency, and energy per inference. These constraints motivate architectures that (i) respect the physical features of radar signatures and sparsity patterns and (ii) reduce arithmetic intensity without sacrificing discriminative Doppler structure.

In this work, we propose RadMamba, a radar-oriented, parameter-efficient Mamba \ac{SSM} for micro-Doppler classification. RadMamba replaces vision-style patching and purely linear token projections with three radar-driven mechanisms: (i) \emph{channel fusion with downsampling} to mix channels and shrink low-information regions; (ii) \emph{Doppler-aligned segmentation} that preserves Doppler–time continuity for state-space modeling; and (iii) \emph{convolutional token projections} to capture Doppler-span variations better and improve inter-patch coherence for the \ac{SSM} backbone.

The main contributions of this work are:
\begin{itemize}
\item A radar-centric state-space architecture that preserves the physical structure of micro-Doppler signatures via Doppler-aligned segmentation and convolutional token projections, reducing $\#$FLOP/Inf. while maintaining temporal–Doppler fidelity.
\item A comprehensive evaluation across \ac{CW} and \ac{FMCW} datasets covering non-continuous and continuous settings: RadMamba attains $99.8\%$ on a dataset measured by the \texttt{\ac{DIAT}}~\cite{DatasetA}, $91.2\%$ on a dataset from the laboratory of \texttt{\ac{CI4R}}~\cite{DatasetB} (competitive with heavier models), and $89.3\%$ on a dataset collected at the \texttt{\ac{UoG20}}~\cite{DatasetC} (at least $3\%$ higher than prior methods with far more parameters), with substantially fewer parameters and lower $\#$FLOP/Inf. across all three datasets.
\item An ablation study showing the synergistic effect of (i) channel fusion with downsampling, (ii) Doppler-aligned segmentation, and (iii) convolutional projections, each contributing to efficiency and accuracy beyond conventional \ac{ViT}/\ac{SSM} patching and linear projections.
\item An open-source PyTorch framework, \texttt{RadHAR}, to facilitate reproducible research and rapid prototyping for micro-Doppler \ac{HAR}.
\end{itemize}

%---------------------------------------------------------------------
\section{Related Works}
\label{sec:related}
%---------------------------------------------------------------------
%---------------------------------------------------------------------
\subsection{Conventional machine learning method}
%---------------------------------------------------------------------

 These methods involved extracting domain-specific features from radar signals, followed by classification using algorithms like \ac{kNN} and \ac{SVM}. For instance, Erol et al.~\cite{Erol2018} used \ac{kNN} to analyze power burst curves in spectrograms for fall detection, distinguishing fall from non-fall events. Li et al.~\cite{Li2019} applied \acp{SVM} to micro-Doppler features such as Doppler centroid, bandwidth, and \ac{SVD} components, while Ding et al.~\cite{Ding2019} extracted 28 features of four types: dynamic Doppler frequency, range change, energy change, and dispersion of range and Doppler from dynamic range-Doppler trajectories for subspace \ac{kNN} classification. Although effective in controlled environments, these approaches struggled to generalize across diverse conditions and activities due to their dependence on hand-crafted features, often leading to overfitting and poor scalability for continuous tasks~\cite{survey2019,Survey-Continuous}.
 
%-------------------------------------------------------------
\subsection{CNN-based method}
%-------------------------------------------------------------
The shift to deep learning marked a turning point for \ac{RadHAR}, with \acp{CNN} leading the charge. Unlike traditional methods, \acp{CNN} automatically learn spatial hierarchies from radar representations like spectrograms and range-Doppler maps. Kang et al.~\cite{Kang2021} employed \acp{CNN} in a two-step process for segmenting and classifying activity sequences, achieving over 97\% segmentation accuracy and 95\% classification accuracy with 60 GHz millimeter-wave radar data. Yu et al.~\cite{Yu2022} introduced a `Dual-View \ac{CNN}' operating on orthogonal projections of voxelized point clouds, reaching 97.61\% accuracy. However, when \acp{CNN} are applied only to micro-Doppler maps treated as images, they lack inherent temporal modeling, which is a critical limitation for continuous activity recognition.

%-------------------------------------------------------------
\subsection{RNN-based method}
%-------------------------------------------------------------

To address temporal dynamics in human activities, researchers have incorporated \acp{RNN}, particularly \ac{LSTM}~\cite{LSTM} and \ac{GRU}~\cite{GRU} networks, which are specifically designed for sequential data processing. Shrestha et al.~\cite{DatasetC} demonstrated the superiority of bidirectional \ac{LSTM} networks over traditional \acp{SVM} and standard LSTMs in continuous activity recognition. Werthen-Brabants et al.~\cite{Bi-RNN} proposed an innovative split \ac{Bi}-RNN architecture for real-time fall alerts, using separate forward \& backward \ac{RNN} components for immediate, refined predictions.

%---------------------------------------------------------------------
\subsection{Hybrid CNN-RNN-based method}
%---------------------------------------------------------------------
Recognizing the complementary strengths of \acp{CNN} and \acp{RNN}, hybrid architectures emerged as a dominant paradigm. These models typically combine \ac{CNN}-based spatial feature extraction with \ac{RNN}-based temporal modeling. Kurtoglu et al.~\cite{CNN-BiLSTM} implemented a sophisticated multi-modal approach, utilizing 3D \acp{CNN} with \ac{Bi}-\ac{LSTM} layers for range-Doppler maps and 2D/1D \acp{CNN} with \ac{Bi}-\ac{LSTM} for spectrograms and envelopes, achieving 93.3\% accuracy for continuous activities mixed with American Sign Language~\cite{wikipedia_asl}. Zhu et al.~\cite{CNN-LSTM} integrated \ac{CNN}-extracted features with \ac{LSTM}-based temporal modeling, achieving 98.65\% accuracy on a non-continuous Doppler radar human activity recognition task with 7 classes. 

For distributed radar systems, hybrid models achieve an accuracy of 87.1\% in continuous activity recognition tasks with only 71\,k parameters~\cite{CNN-RNN-distributed}. Compared to \ac{CNN}-based models in non-continuous tasks that utilize 135\,M parameters~\cite{Tang2021} and \ac{RNN}-based models in continuous tasks that utilize 189\,M parameters~\cite{DatasetC} to achieve their best classification accuracy, hybrid models are more lightweight. However, the $\#$FLOP/Inf. required by the hybrid model~\cite{CNN-RNN-distributed} is evaluated to be at least 1\,G. The arithmetic intensity of \acp{CNN}, especially for multi-channel signals, drives their computational complexity.

%---------------------------------------------------------------------
\subsection{Transformer and SSM-based method}
%---------------------------------------------------------------------

Recently, researchers have explored transformer-based architectures and \ac{SSM} in \ac{RadHAR}. \ac{ViT}-based methods~\cite{FML-VIT} achieved 92\% accuracy with 2.7\,M parameters, while~\cite{LW_VIT} reached 92.1\% with 769\,k parameters on non-continuous \ac{FMCW} radar data~\cite{DatasetA}, though both demand significant $\#$FLOP per inference. (169\,M and 2.41\,G, respectively). In continuous \ac{HAR}, however, \ac{ViT} appears to underperform conventional \ac{CNN}-based models, such as the first three blocks of AlexNet and ResNet~\cite{VitAIHAR2022}. SSM-based approaches~\cite{ActivityMamba} hit 99.82\% accuracy on a non-continuous \ac{CW} dataset~\cite{DatasetA} with 8.7\,M parameters and 1.22G $\#$FLOP per inference. Compared to ultra-lightweight object detection models optimized for edge devices, such as YOLO-Fastest with 350\,k parameters, and 252\,M \#FLOP/Inf, directly applying these advanced architectures to radar-based HAR tasks remains costly due to their high computational overhead.

%----------------------------------------------
\section{Proposed RadMamba}
\label{sec:radmamba}
\begin{figure*}[!t]
    \centering
    \includegraphics[width=1\linewidth]{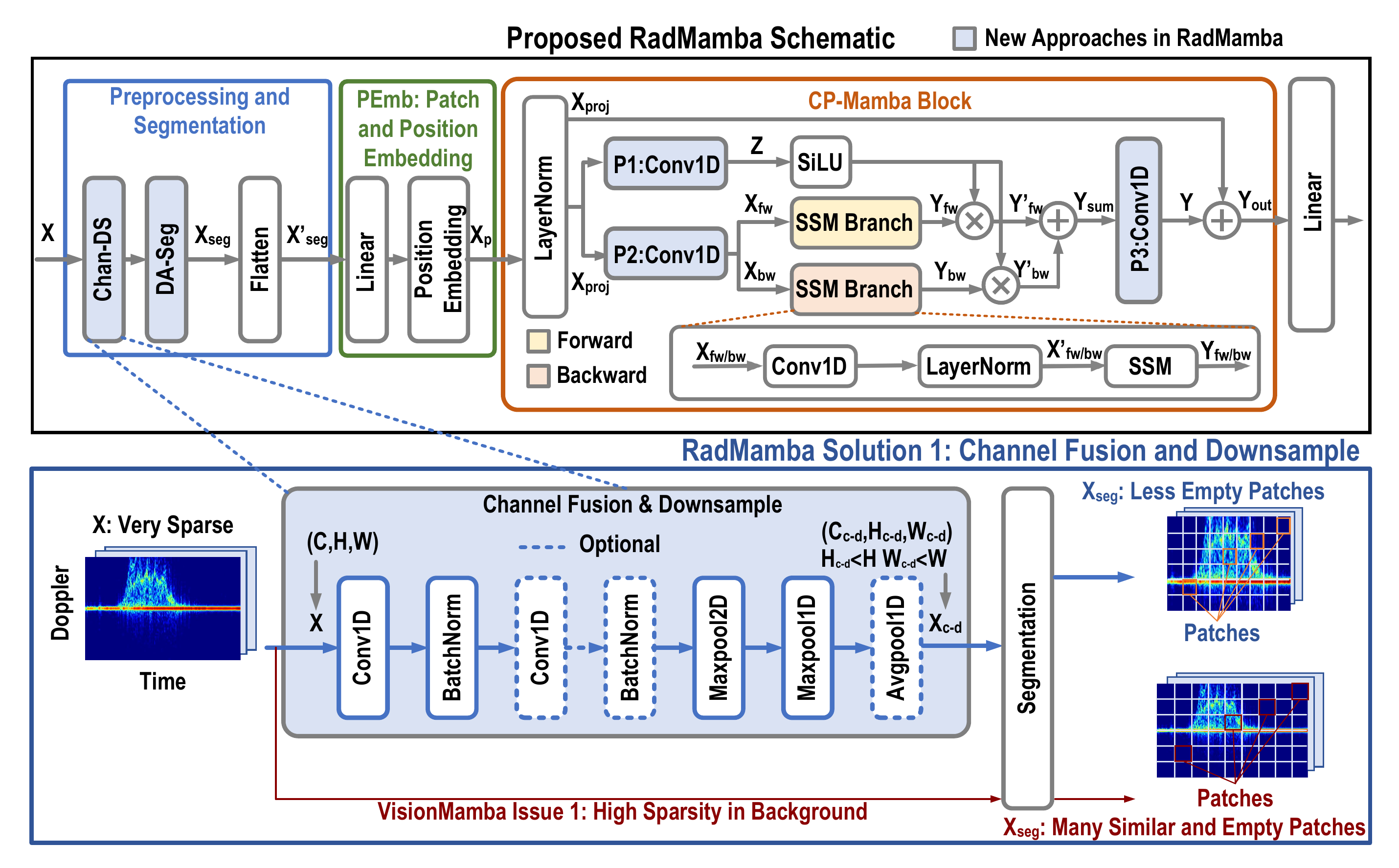}
    \caption{The schematic of the proposed RadMamba architecture and its novel Solution 1 for radar micro-Doppler-based classification.}
    \label{fig:MOD}
\end{figure*}

%----------------------------------------------
To effectively combine the modeling power of \ac{SSM} with lower computational cost, it is essential to exploit the distinctive characteristics of micro-Doppler signatures. However, applying a conventional \ac{ViM} to \ac{RadHAR} reveals three key issues:
\begin{itemize}
    \item \textbf{Issue 1:} Micro-Doppler signatures exhibit significant background sparsity, so most time–frequency bins carry little task-relevant information.
    \item \textbf{Issue 2:} Rectangular patch segmentation disrupts the continuous spatio-temporal structure of micro-Doppler patterns, degrading representation quality.
    \item \textbf{Issue 3:} Standard linear projection layers struggle to capture Doppler-specific, time–frequency patterns.
\end{itemize}
To clarify these points, we trace the data flow from input to \ac{SSM} in RadMamba, highlighting issues specific to radar micro-Doppler-based \ac{HAR} tasks and describing our proposed solutions in RadMamba.  As shown in Fig.~\ref{fig:MOD}, RadMamba consists of three core components: preprocessing and segmentation; patch and positional embedding; and a convolution-projection–based Mamba block with bidirectional \acp{SSM}. We customize the architecture for micro-Doppler signals using three solutions: channel fusion with downsampling, Doppler-aligned segmentation, and the \ac{CP-Mamba} block.

%---------------------------------------------
\subsection{Solution 1: Channel Fusion and Downsampling}
\label{subseq:CD}
%---------------------------------------------

The micro-Doppler signature $\mathbf{X} \in \mathbb{R}^{C \times H \times W}$ serves as input to the classifier, with the Doppler frequency dimension represented as the image height $H$ and time as the image width $W$. The channel dimension $C$ corresponds either to multiple radar signal channels or to RGB channels when visualized as an image. In the subsequent \textit{Segment} step, $\mathbf{X}$ is partitioned into a matrix of smaller patches, denoted as $\mathbf{X}_{seg}$. As depicted in Fig.\ref{fig:MOD} VisionMamba Issue 1, many patches in $\mathbf{X}_{seg}$ contain only background information and therefore appear visually redundant. Based on our analysis of the CI4R Dataset~\cite{DatasetB}, the micro-Doppler spectrograms exhibit an average spectrogram sparsity of approximately 87\%. Such sparse background regions contribute limited discriminative information regarding Doppler frequency or temporal dynamics, reducing the representational effectiveness of these patches.

To address the high sparsity inherent in micro-Doppler signatures, we introduce a preprocessing pipeline, Channel Fusion and Downsampling (Chan-DS), before the segmentation stage, as depicted in Fig.~\ref{fig:MOD}.
\begin{align}
    \mathbf{X_{c-d} = \text{Chan-DS}(\mathbf{X})}
\end{align}

where $\mathbf{X}_{c-d} \in \mathbb{R}^{C_{c-d} \times H_{c-d} \times W_{c-d}}$ with $H_{c-d}<H$ and $W_{c-d}<W$ is the micro-Doppler signature after channel fusion and downsampling.
The architecture of the \textit{Chan-DS} block is as follows:
\begin{align}
    \text{Chan-DS} = [&(\text{Conv2D}, \text{BatchNorm}) \times L, \nonumber\\ 
    &\text{Maxpool2D},\nonumber\\
    &\text{Maxpool1D},\nonumber\\
    &\text{Avgpool1D}\text{ (optional)}]\nonumber
\end{align}
where $L = 1,2$. The \ac{2D} convolutional layer and batch normalization layer mix channel information and extract preliminary features from the input tensor $\mathbf{X}$. A \ac{2D} max pooling layer and a \ac{1D} max pooling layer reduce spatial dimensions. Optionally, an additional \ac{1D} average pooling layer further downsamples the time dimension if the data sparsity along the time axis is excessive for \ac{FMCW} radar datasets.

This approach avoids generating numerous empty or redundant patches. Additionally, downsampling reduces the spatial dimensions before segmentation, thereby further contributing to an energy-efficient design.

%----------------------------------------------
\subsection{Solution 2: Doppler-Aligned Segmentation and Patch Embedding}
\label{subseq:DAS}
%----------------------------------------------

\begin{figure*}
    \centering
    \includegraphics[width=\linewidth]{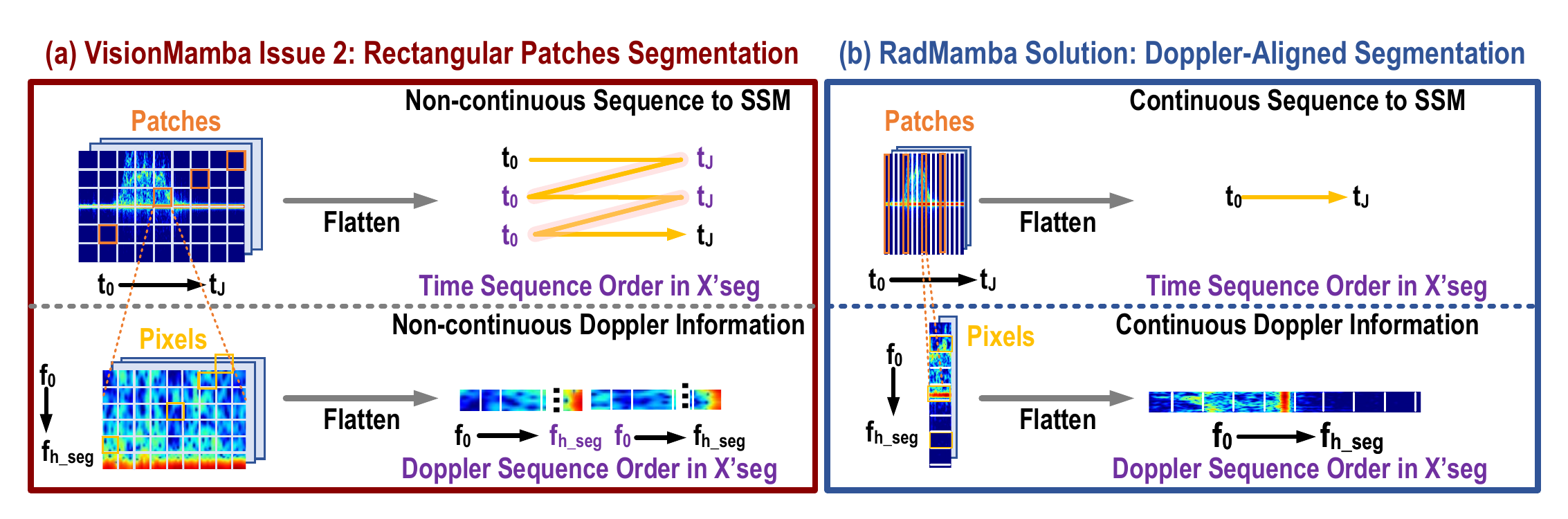}
    \caption{VisionMamba Issue 2 and RadMamba Solution 2: Doppler-Aligned Segmentation}
    \label{fig:DASeg}
\end{figure*}
The feature map after segmentation $\mathbf{X}_{seg}$  has dimensions of  $C \times I \times J \times H_{seg} \times W_{seg}$, where $H_{seg}$ and $W_{seg}$ denote the height and width of each patch. Thus, the number of patches per column and per row are determined by $I = \frac{H}{H_{seg}}$ and $J = \frac{W}{W_{seg}}$, respectively, resulting in a total number of patches $N = IJ$.

As illustrated in Fig.~\ref{fig:DASeg} (a) (upper), patches indexed by $j$ along the horizontal direction correspond to temporal positions $t_j$, where $j = 0, 1, \ldots, J$. Within each patch, as shown in Fig.~\ref{fig:DASeg} (a) (bottom), pixels indexed by $h_{seg}$ along the vertical direction correspond to Doppler frequency points $f_{h_{seg}}$, where $h_{seg} = 0, 1, \ldots, H_{seg}$. Each patch is then flattened into a vector:
\begin{align}
\mathbf{X^{\prime}_{seg}} = \text{Flatten}(\mathbf{X_{seg}})
\end{align}
where $\mathbf{X^{\prime}_{seg}} \in \mathbb{R}^{N \times CH_{seg}W_{seg}}$. 

The second limitation of conventional VisionMamba lies in the fact that patch flattening disrupts Doppler–temporal continuity, as depicted in Fig.~\ref{fig:DASeg} (a). On the one hand, rectangular segmentation results in a temporal ordering in $\mathbf{X}^{\prime}_{seg}$ along the dimension $N$ of the form:
\begin{align}
    [t_0, t_1, \cdots, t_J, t_0, t_1, \cdots, t_J, \cdots]
\end{align}represented by the yellow folded arrow in Fig.~\ref{fig:DASeg} (a). Since VisionMamba treats $N$ as a temporal evolution dimension in the \ac{SSM} branches, the artificial jump from $t_J$ to $t_0$ introduces a temporal discontinuity, making it difficult for the \ac{SSM} to extract meaningful temporal dynamics.
On the other hand, the flattening operation yields a Doppler frequency sequence inside each patch of the form:
\begin{align}
    [f_0, f_1, \cdots, f_{h_{seg}}, f_0, f_1, \cdots, f_{h_{seg}}, \cdots]
\end{align}as shown in Fig.~\ref{fig:DASeg} bottom. The discontinuity between $f_{h_{seg}}$ and $f_0$ breaks the frequency pattern within each patch, simultaneously degrading both Doppler frequency consistency and the temporal evolution represented in that patch.

To address this issue, we propose Doppler-aligned segmentation, as shown in Fig.~\ref{fig:DASeg} (b): 
\begin{align}
        \mathbf{X_{seg}} = \text{DA-Seg}(\mathbf{X}_{c-d})
\end{align}
Specifically, we define patches with a height of $H_{seg} = H_{c-d}$ (the full Doppler dimension) and a width of $W_{seg} = 1$ (a single time bin of the spectrogram). 

This Doppler-aligned segmentation ensures that the temporal evolution of the patch sequence remains strictly ordered as:
\begin{align}
    [t_0, t_1, \cdots, t_J]
\end{align}corresponding directly to the time sequence $W$. When the segmented data is fed into the \ac{SSM}, as shown in Fig.~\ref{fig:DASeg} (b) (upper), the \ac{SSM} receives a continuous temporal input stream. Furthermore, Doppler-aligned segmentation produces a patch vector with dimension
\begin{align}
    CH_{seg}W_{seg}= H_{c-d}
\end{align}
preserving the entire Doppler profile within each patch. Consequently, after flattening, the Doppler frequency sequence remains intact as:
\begin{align}
    [f_0, f_1, \cdots, f_{h\_seg}]
\end{align}as illustrated in Fig.~\ref{fig:DASeg}(b) (bottom). This maintains Doppler continuity, enabling the model to leverage fine-grained frequency dynamics without disruption.
%----------------------------------------------
\subsection{Solution 3: Convolution Projections in CP-Mamba Blocks}
\label{subseq:proj}
%----------------------------------------------
Within the conventional Mamba block (Fig.~\ref{fig:MOD}), the input $\mathbf{X}_{p}$ is first normalized to obtain $\mathbf{X}_{proj}$, which is then split into three paths:
\begin{align}
\mathbf{Z} &= \text{Projection}_{P1}(\mathbf{X}_{proj}), \\
\mathbf{X}_{fw} &= \text{Projection}_{P2}(\mathbf{X}_{proj}),\\
\mathbf{X}_{bw} &= \text{Flip}(\mathbf{X}_{fw},dim = \text{temporal dim.}),
\end{align}
where $\mathbf{Z}, \mathbf{X}_{fw}, \mathbf{X}_{bw}  \in \mathbb{R}^{N\times dim}$, and Flip() reverses the temporal dimension of $\mathbf{X}_{fw}$. 

In the conventional Mamba block, $\text{Projection}_{P1}$ and $\text{Projection}_{P2}$ are linear layers. In contrast, the \ac{CP-Mamba} block replaces them with 1D convolutions using kernel size $(dim, dim, 3)$. Next, $\mathbf{X}_{fw}$ and $\mathbf{X}_{bw}$ are processed by the two \ac{SSM} branches:
\begin{align}
[\text{Conv1D}{fw/bw}, \text{LayerNorm}{fw/bw}, \text{SSM}{fw/bw}],
\end{align}
where $\text{Conv1D}{fw/bw}$ has weight shape $(dim, dim, 1)$ and operates along the sequence dimension $(N, dim)$.

The two directional outputs are gated by the \ac{SiLU}-activated $\mathbf{Z}$ and summed:
\begin{align}
\mathbf{Y}'{fw/bw} &= \mathbf{Y}{fw/bw} \odot \text{SiLU}(\mathbf{Z}), \\
\mathbf{Y}{sum} &= \mathbf{Y}'{fw} + \mathbf{Y}'{bw},
\end{align}
where $\odot$ denotes element-wise multiplication. Finally, a projection $\textit{P3}$ and residual connection produce the output:
\begin{align}
\mathbf{Y}_{out} = \text{Projection}_{P3}(\mathbf{Y}_{sum}) + \mathbf{X}_{proj},
\end{align}

In the conventional Mamba block, the projection layer $\textit{P3}$ is linear. In our \textit{\ac{CP-Mamba}}, $\textit{P3}$ is replaced with a $\text{Conv1D}$, where $\text{Conv1D}_{P3}$ has weights $(dim, dim, 1)$ and restores the sequence shape to $(N, dim)$.

\begin{table}[!t]
\centering
\caption{Averaged cross-correlation between
patches before and after each projection layer.}
\label{tab:correlation}
\resizebox{\linewidth}{!}{%
\begin{tabular}{|ccccc|}
\hline
\multicolumn{1}{|c|}{} &
  \multicolumn{2}{c|}{\begin{tabular}[c]{@{}c@{}}Linear Projection Layers\\ in Conventional Mamba\end{tabular}} &
  \multicolumn{2}{c|}{\begin{tabular}[c]{@{}c@{}}Convolutional Projection Layers\\ in Our Proposed \ac{CP-Mamba}\end{tabular}} \\ \hline \hline
\multicolumn{1}{|c|}{Projection} &
  \multicolumn{1}{c|}{At Input} &
  \multicolumn{1}{c|}{At Output} &
  \multicolumn{1}{c|}{At Input} &
  At Output \\ \hline
\multicolumn{1}{|c|}{1} & \multicolumn{1}{c|}{0.008} & \multicolumn{1}{c|}{77.9} & \multicolumn{1}{c|}{0.002} & 1065.7 \\ \hline
\multicolumn{1}{|c|}{2} & \multicolumn{1}{c|}{0.008} & \multicolumn{1}{c|}{0.4}  & \multicolumn{1}{c|}{0.002} & 16.1   \\ \hline
\multicolumn{1}{|c|}{3} & \multicolumn{1}{c|}{160.1} & \multicolumn{1}{c|}{75.5} & \multicolumn{1}{c|}{245.3} & 356.4  \\ \hline
\end{tabular}%
}
\end{table}

Ideally, projection layers should improve cross-correlation between patches, since a strong classifier should treat Doppler-aligned patches of the same activity as a coherent unit. Linear projections, however, increase correlation only marginally (P1, P2) or even reduce it (P3), limiting classification performance. The convolutional projections in \ac{CP-Mamba} consistently strengthen cross-correlation, making patches easier to identify as belonging to the same action and improving Doppler-pattern sensitivity.

Tab.~\ref{tab:correlation} reports the averaged patch cross-correlation using the Dataset \texttt{\ac{CI4R}}~\cite{DatasetB}. For conventional linear projections, the averaged correlation before projections 1 and 2 is 0.008, rising to 0.77 and 0.4, respectively. Before and after projection 3, it drops from 160.1 to 75.5. In contrast, our convolutional projections greatly enlarge the correlation gap (e.g., from 77.9 to 1065.7 for P1), indicating superior ability to aggregate time bins of the same activity.

To compute the averaged patch cross-correlation, we train the model, freeze the parameters, and refeed the data. For $\mathbf{X}_{proj} \in \mathbb{R}^{B \times N \times D}$, each patch is indexed by $n$. The averaged cross-correlation is defined as:
\begin{align}
Corr_{avg} = &\frac{1}{B} (\frac{1}{N}\sum_{n = 1}^{N}(\frac{1}{N}\sum_{n^{\prime} = 1}^{N}( \nonumber\\
& corr(\mathbf{X_{proj}(b, n)} , \mathbf{X_{proj}(b, n^{\prime})}))))
\end{align}
where the correlation between two vectors of equal length is computed as:
\begin{align}
&corr(\mathbf{A} , \mathbf{B})=  \sum_{m=0}^{M}\sum_{m^{\prime}=0}^{M} \mathbf{A}[m] \cdot \mathbf{B}[m+m^{\prime}]
\end{align}
%----------------------------------------------
\subsection{Selective State-Space Modeling for Micro-Doppler Sequences}
\label{subseq:SSM}
%----------------------------------------------
\begin{figure}[!t]
    \centering
    \includegraphics[width=\linewidth]{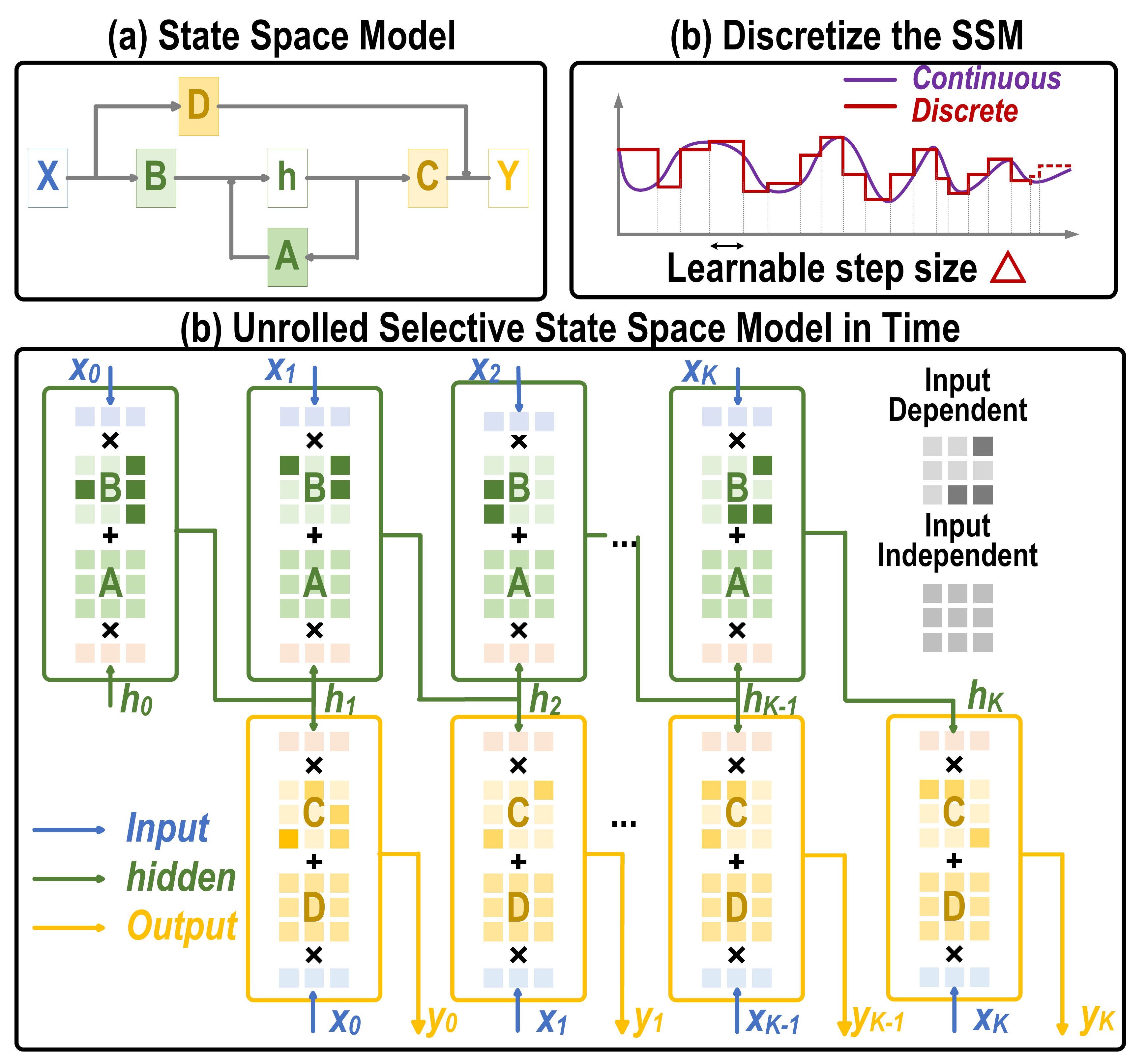}
    \caption{The architecture of the State Space Model in RadMamba.}
    \label{fig:SSM}
\end{figure}
The \ac{SSM} is a significant layer in \ac{CP-Mamba} that enables efficient sequence modeling for the dynamic nature of micro-Doppler signatures. As shown in Fig.~\ref{fig:SSM} (a), the discrete \ac{SSM} is defined as: 
\begin{align}
\mathbf{h}[n] &= {\mathbf{\bar A}}[n] \mathbf{h}[n-1] + {\mathbf{\bar B}}[n] \mathbf{x}[n], \\
\mathbf{y}[n] &= \mathbf{C}\mathbf{h}[n] + \mathbf{D}\mathbf{x}[n],
\end{align}

where the state transition ${\mathbf{\bar A}}$ and the input matrices dimension ${\mathbf{\bar B}}$ are discretized from corresponding matrices ${\mathbf{A}}$ and ${\mathbf{B}}$ in continuous \ac{SSM} by a learnable step size $\Delta$ as shown in Fig.~\ref{fig:SSM} (b):
\begin{align}
\bar{\mathbf{A}} &= \exp(\boldsymbol{\Delta} \cdot \mathbf{A}), \\
\bar{\mathbf{B}} &= \left( \int_0^{\boldsymbol{\Delta}} \exp(\mathbf{A}\tau) \, d\tau \right) \mathbf{B},
\end{align}

The input of the SSM branches in the proposed \ac{CP-Mamba} block are tensors $\mathbf{X}_{fw/bw} \in \mathbb{R}^{N\times dim}$; after \textit{Conv1D} and \textit{LayerNorm} in SSM branch, this is termed as $\mathbf{X'}_{fw/bw} \in \mathbb{R}^{N\times dim}$.

Therefore, the discretized state-transition tensor ${\mathbf{\bar A}}$ has dimension $N \times dim \times dim_{s}$, and the input tensor ${\mathbf{\bar B}}$ has dimension $N \times N \times dim_{s}$. Here, $N$ equals the time dimension in the micro-Doppler signatures. The input tensor $\mathbf{\bar B}$ captures how the Doppler-aligned input vector influences the state $\mathbf{h}[n]$, whereas the state-transition tensor ${\mathbf{\bar A}}$ captures how the previous state $\mathbf{h}[n-1]$ contributes to $\mathbf{h}[n]$.
Moreover, the matrices $\mathbf{C} \in \mathbb{R}^{N \times dim_{s}}$ translate the current state $\mathbf{h}[n]$ to the output. The skip matrices $\mathbf{D} \in \mathbb{R}^{dim}$ add weighted input directly to the output of \ac{SSM}, which helps the recurrent part focus on the temporal evolution.  

The selective mechanism in the \ac{SSM} introduces input-dependent parameterization that enhances adaptability as shown in Fig.~\ref{fig:SSM} (c). The parameters $\mathbf{B}$, $\mathbf{C}$, and $\boldsymbol{\Delta}$ are dynamically generated from the input sequence via linear projections:
\begin{align}
\mathbf{B} &= \mathbf{W}_{\mathbf{B}}\mathbf{X}, \\
\mathbf{C} &= \mathbf{W}_{\mathbf{C}}\mathbf{X}, \\
\boldsymbol{\Delta} &= \text{Softplus}(\mathbf{W}_{\boldsymbol{\Delta}}\mathbf{X}),
\end{align}

where $\mathbf{W}_{\mathbf{B}}$ and $\mathbf{W}_{\mathbf{C}}$ are weights of dimension $(dim, dim_{s})$. The $\boldsymbol{\Delta}$ computation involves a linear layer with weights $(dim, \text{dt\_rank})$ followed by a projection to $(N, dim)$ via weights $(\text{dt\_rank}, dim)$, where $\text{dt\_rank}$ is a hyperparameter controlling the rank of the dynamic tensor. The Softplus activation ensures positive step sizes, implemented as $\text{Softplus}(x) = \log(1 + \exp(x))$.

%----------------------------------------------
\subsection{Overall Architecture}
%----------------------------------------------
To summarize the method, the proposed RadMamba pipeline in pseudo-code is presented in Algorithm I. The patch embedding in \textit{PEmb} follows the method in ~\cite{Vim}, and the position embedding method is a sinusoidal position encoding.~\cite{PEmb}.

\begin{algorithm}[!t] 
\caption{Proposed RadMamba Pipeline}
\textbf{Input:} Micro-Doppler Signature $\mathbf{X} \in \mathbb{R}^{C \times H \times W}$\\
\textbf{Parameters:} $dim$, $dim_{s}$, $dt\_rank$\\
\textbf{Output:} Classification Result $\mathbf{F} \in \mathbb{R}^{Q}$, where \(Q\) is the number of classes
\begin{algorithmic}[1]
\State \textbf{Channel Fusion and Downsample:}
\State \hspace{1em}$\mathbf{X_{c-d}} = \textit{Chan-DS}(\mathbf{X})$, $\mathbf{X}_{c-d} \in \mathbb{R}^{C_{c-d} \times H_{c-d} \times W_{c-d}}$

\State \textbf{Doppler-Aligned Segmentation:}
\State \hspace{1em}$\mathbf{X_{seg}} = \textit{DA-Seg}(\mathbf{X}_{c-d})$, $\mathbf{X}_{seg} \in \mathbb{R}^{C_{c-d} \times N\times H_{c-d}}$

\State \textbf{Patch and Position Embedding:}
\State \hspace{1em}$\mathbf{X'}_{p} = \textit{PEmb}(\mathbf{X}_{seg})$, $\mathbf{X'}_{p} \in \mathbb{R}^{N \times D}$

\State \textbf{\ac{CP-Mamba} Block:} 
\State \hspace{1em}$\mathbf{Y}_{out} = \textit{\ac{CP-Mamba}}(\mathbf{X}_{p})$, $\mathbf{Y}_{out} \in \mathbb{R}^{N\times dim}$

\State \textbf{Classify:}
\State \hspace{1em}$\mathbf{F} = \mathbf{W}_{out}\mathbf{Y}_{out}$, $\mathbf{F} \in \mathbb{R}^{Q}$
\end{algorithmic}
\end{algorithm}

%--------------------------------------------------
\section{Open-source Datasets and Preprocessing for Radar-based HAR}
\label{sec:dataset}
\begin{figure}[!t]
    \centering
    \includegraphics[width=\linewidth]{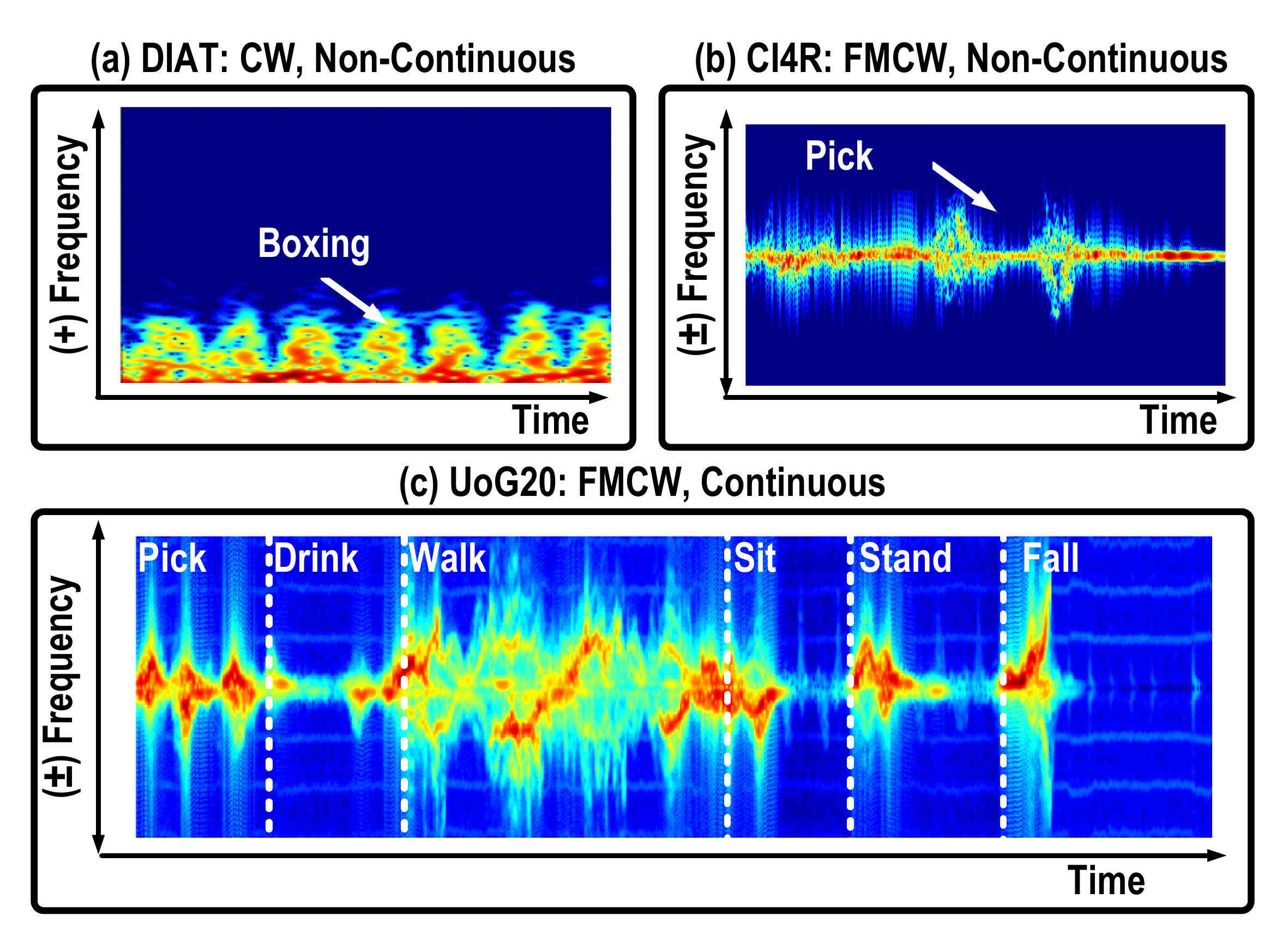}
    \caption{Micro-Doppler Signature Example of (a) Dataset DIAT (b) Dataset CI4R (c) Dataset UoG20}
    \label{fig:MD_example}
\end{figure}

%--------------------------------------------------
To evaluate the efficacy of our proposed RadMamba framework, we utilize three distinct radar datasets. This section describes these datasets, namely the \ac{CW} non-continuous Dataset \texttt{\ac{DIAT}}, \ac{FMCW} non-continuous Dataset \texttt{\ac{CI4R}}, and \ac{FMCW} continuous Dataset \texttt{\ac{UoG20}}, along with their respective pre-processing steps. More details on these datasets are summarized in Tab.~\ref{tab:data}. Here, the word `continuous' refers to samples of data where sequences of human activities are continuously performed one after the other. Examples of micro-doppler signature after pre-processing are shown in Fig.~\ref{fig:MD_example}.

\begin{table}[!t]
\caption{Overview of radar-based human activity recognition datasets used in this work.}
\label{tab:data}
\resizebox{\linewidth}{!}{%
\begin{tabular}{|cccc|}
\hline
Datasets & \texttt{\ac{DIAT}}~\cite{DatasetA} & \texttt{\ac{CI4R}}~\cite{DatasetB} & \texttt{\ac{UoG20}}~\cite{DatasetC} \\ \hline \hline
Continuous & No & No & Yes \\
Classes & 6 & 11 & 6 \\
Sensor & X-band \ac{CW} & IWR1443 \ac{FMCW} & \ac{FMCW} \\
Freq.(GHz) & 10 & 77 & 5.8 \\
Resolution & 224$\times$224 & 224$\times$224 & 240$\times$1471 \\
BW(MHz) & N/A & 750 & 400 \\
Length per sample (s) & 3 & 20 & 35 \\
Train Samples & 2646 & 528 & 39 \\
Test Samples & 759 & 132 & 6 \\
Sliding Window & N/A & N/A & 224 \\\hline
\end{tabular}%
}
\end{table}
%-------------------------------------------------
\subsection{Dataset DIAT: X-Band CW Radar with Discrete Activity Instances~\cite{DatasetA}}
\label{subsec:datasetA}
%-------------------------------------------------
This dataset was collected using an X-band \ac{CW} radar operating at a frequency of 10 GHz. It encompasses data from 30 human subjects performing `suspicious' activities at ranges spanning 10 meters to $\pm$0.5 kilometers over 3-second durations, resulting in an imbalanced set of 3,780 spectrogram samples of shape  $(3, 224, 224)$ across six activity classes. Preprocessing adheres to the methodology outlined in~\cite{DatasetA}. To facilitate a fair comparison with previous works, the dataset is partitioned into 70\% training, 10\% validation, and 20\% test sets within each class, maintaining class-specific balance for reliable performance evaluation.

\subsection{Dataset CI4R: Short-Range FMCW Radar for Multi-Class Activity Recognition~\cite{DatasetB}}
\label{subsec:datasetB}
For the second dataset, experiments employ a Texas Instruments IWR1443 \ac{FMCW} radar, operating at 77 GHz with a bandwidth of 750 MHz. It includes data from six participants, varying in age, height, and weight, performing 11 distinct activities and ambulatory gaits. Each participant executed 10 repetitions per activity, yielding 60 samples per class per sensor, as detailed in~\cite{DatasetB}. Preprocessing involves applying a 256-point short-time Fourier transform (\ac{STFT}) with a context window length of 256 to convert the raw radar signals into a \ac{2D} micro-Doppler representation of shape $(1, 224, 224)$. The dataset is split with an 8:2 training-to-test ratio, ensuring sufficient data for model optimization and evaluation.

\subsection{Dataset UoG2020: Continuous FMCW Radar for Realistic HAR Sequences~\cite{DatasetC}}
\label{subsec:datasetC}
Representing a leap in complexity, this dataset was acquired using an \ac{FMCW} radar operating at 5.8 GHz with a 400 MHz bandwidth. It comprises data from 15 participants (14 males, 1 female, aged 21–35) performing six basic activities within continuous 35-second sequences. These activities are organized into three distinct sequential orders, with participants granted autonomy in selecting transition points between consecutive activities and determining trajectories for translational tasks, as described in~\cite{DatasetC}. 
The resulting micro-Doppler signatures are preprocessed into $(1, 240, 224)$ \ac{2D} frames, with data from two participants designated as the test set and data from the remaining 13 participants used for training, providing a challenging benchmark for continuous sequence modeling.

%--------------------------------------------------
\section{Experimental Setup}
\label{sec:experifments}
%--------------------------------------------------
%--------------------------------------------------
\subsection{Network Structure}
%--------------------------------------------------

\begin{table}[!t]
\caption{Network Architecture of Models Used in This Work}
\label{tab:network}
\resizebox{\linewidth}{!}{%
\begin{tabular}{|cc|}
\hline
Model & Swept Hidden Size ($dim$ for RadMamba) \\ \hline \hline
ResNet6 (Customized) & 5, 7, 12, 20, 30 \\
\ac{Bi}-\ac{LSTM} & 1, 2, 4, 10, 20, 28, 34 \\
\ac{CNN}-\ac{LSTM} & 2, 4, 6, 16, 28, 34, 46 \\
\ac{CNN}-\ac{Bi}-\ac{GRU} & 1, 2, 4, 6, 16, 28 \\ 
RadMamba Dataset \texttt{\ac{DIAT}} & 8, 16, 32, 64, 80 \\
RadMamba Dataset \texttt{\ac{CI4R}} & 8, 16, 32, 64, 80, 96, 128, 160 \\
RadMamba Dataset \texttt{UoG20} & 8, 16, 20, 24, 32  \\\hline 
\end{tabular}%
}
\end{table}

The classification performance of RadMamba is evaluated against a suite of established models, selected for their proven efficacy across \ac{CNN}, \ac{RNN}, and hybrid architectures: customized ResNet6~\cite{ResNet}, \ac{Bi}-\ac{LSTM}~\cite{DatasetC}, \ac{CNN}-\ac{LSTM}~\cite{CNN-LSTM}, and \ac{CNN}-\ac{Bi}-\ac{GRU}~\cite{CNN-RNN-distributed}. To explore the impact of model scaling, we conducted a model size sweep experiment, varying the hidden dimensions as shown in Tab.~\ref{tab:network}. These sizes are chosen to capture the full spectrum of performance trends as model capacity increases, providing insights into scalability and efficiency. 

The customized ResNet6 is derived from ResNet18 by preserving the first five layers and the final classifier while maintaining its channel ratio between consecutive channels. The minimum channel width is tunable to target low-FLOP regimes, enabling fair comparisons under limited computation conditions. Although the \ac{CNN}-\ac{Bi}-\ac{GRU} retains a larger parameter budget, its fusion strategy may not generalize consistently across datasets.

All four baselines (ResNet6, Bi-LSTM, CNN-LSTM, and CNN-Bi-GRU) are retrained from scratch in a unified PyTorch pipeline, using consistent dataset splits, preprocessing, optimization settings, and batch=1 memory profiling. For transformer and \ac{SSM} competitors, FML-ViT~\cite{FML-VIT}, LW-ViT~\cite{LW_VIT}, and ActivityMamba~\cite{ActivityMamba}, we report their best published small-scale configurations. Following their protocols, we adopt the same training settings for FML-ViT and ActivityMamba, but use a different input size than LW-ViT: their preprocessing produces 656×656 inputs on the CI4R dataset, whereas our pipeline uses 224×224.

\begin{table}[!t]
\centering
\caption{Specified Network Architecture Variations of RadMamba}
\label{tab:radmambaset}
\resizebox{\linewidth}{!}{%
\begin{tabular}{|c|cccc|}
\hline
Datasets  & $dim_{s}$ & $dt\_rank$ & Chan. Fusion & DS Redu. Factor \\ \hline \hline
Dataset \texttt{\ac{DIAT}}  & 1 & 2 & 2, (16, 3, 3)& (2, 2) \\
Dataset \texttt{\ac{CI4R}}  & 4 & 0 & 1, (1, 3, 3) & (2, 8) \\
Dataset \texttt{\ac{UoG20}} & 16 & 4 & 1, (1, 3, 3) & (2, 32) \\ \hline 
\end{tabular}%
}
\end{table}
For RadMamba, the configurations of channel fusion, downsampling, and \ac{SSM} parameters are dataset-specific to optimize micro-Doppler processing (Tab.~\ref{tab:radmambaset}). The `channel fusion' column specifies [L, ($C_{c-d}, H_k, W_k$)], where L is the number of channel fusion blocks and the $C_{c-d}, H_k, W_k$ are channel, kernel height, kernel width of Conv2D layer, respectively. The `DS Redu. Factor' denotes the reduction factors ($H/H_{c-d}$, $W/W_{c-d}$).

Hyperparameter selection follows two principles: 
\begin{itemize}
    \item \textbf{Bayesian optimization:}  
    Because RadMamba is highly configurable, we perform a Bayesian-optimization-based search over a constrained space. Since $dim$ predominantly determines the sizes of most weight matrices, the number of channels in \textit{Chan-DS}, $dim_s$, and $dt_{rank}$ were selected to minimize model complexity across different $dim$.
    \item \textbf{Physical perspective from micro-Doppler signature (Temporal downsampling and Reduction Factor):}  
    The first downsampling factor of $2\times2$ in \textit{Chan-DS} matches the fact that most Doppler energy appears in the central part of the spectrum. 
    % For CW radar, micro-Doppler remains continuously embedded in time, thus temporal reduction is limited to $2\times$ for the \textit{DIAT} datasets. In FMCW radar, Doppler is derived from phase differences across chirps, resulting in a sparser temporal representation compared to CW radar.
    With horizontal reduction factors of $8$ and $32$ for the \textit{CI4R} and \textit{UoG20} datasets (as shown in Tab.~\ref{tab:radmambaset}), each Doppler vector spans approximately 640\, ms. Empirically, this temporal interval provides an optimal balance for both FMCW datasets.
\end{itemize}
%--------------------------------------------------
\subsection{Training Hyperparameter}
%--------------------------------------------------
Training is conducted using the \ac{AdamW} optimizer~\cite{kingma2014adam,loshchilov2018decoupled}, paired with the \texttt{ReduceLROnPlateau} scheduler, which adjusts the learning rate adaptively based on test performance. For Dataset \texttt{\ac{DIAT}}, the initial learning rate is set to $1 \times 10^{-4}$, with a batch size of 16, and training proceeds for 50 epochs. For Dataset \texttt{\ac{CI4R}}, RadMamba employs an initial learning rate of $5 \times 10^{-3}$, while the baseline models use $1 \times 10^{-4}$, with a batch size of 16 over 100 epochs, reflecting the dataset's increased complexity and sample size. For Dataset \texttt{
\ac{UoG20}}, the initial learning rate is $5 \times 10^{-5}$, with a larger batch size of 256 and 50 epochs, processing continuous sequences with a frame length of 224 and a stride of 1 to effectively capture temporal dependencies. All experiments are executed on a single NVIDIA RTX 4090 GPU.

To account for training variability, all models, including RadMamba, are trained and evaluated ten times with distinct seeds. The following analysis reports the mean and standard deviation of performance metrics from the model size swept experiment.

\subsection{Evaluation Settings}
Inference time and memory profiling are conducted on an NVIDIA Jetson Orin Nano Developer Kit with PyTorch 2.8.0 and CUDA 12.6. The kit is equipped with a 6-core Arm Cortex-A78AE CPU and an NVIDIA Ampere Architecture GPU with unified off-chip 8\,GB 128-bit LPDDR5 memory. The evaluation follows a standardized protocol: each model undergoes 10 warm-up iterations followed by 100 measurement iterations, with inference performed at a batch size of 1. Peak memory (Peak Mem.) is measured using \texttt{torch.cuda.memory\_allocated(device)}, which returns the maximum allocated memory across all 100 inference iterations. CUDA synchronization with the ARM CPU is enforced before and after each forward pass to ensure accurate timing measurements.
%------------------------------------------------------------------------
\section{Results and Discussion}
\label{sec:results}
%-------------------------------------------------------------------------
\subsection{Comparison with Previous Works}
\begin{table*}[!t]
    \caption{Mean and Standard Deviation of Classification Accuracy Performance across Seed 0 to 9 of Different NN-based \ac{RadHAR} Models Evaluated With Three Different Radar Settings alongside Their Model Size and Floating-point Operations ($\#$FLOP.)}
    \label{tab:model_comp}
    \large
    \resizebox{\linewidth}{!}{%
    \begin{threeparttable}
    \begin{tabular}{|l|ccccc|ccccc|ccccc|}
    \hline \hline
    \multicolumn{1}{|c|}{} & \multicolumn{5}{c|}{\begin{tabular}[c]{@{}c@{}}Dataset \texttt{\ac{DIAT}}~\cite{DatasetA}\\ (\ac{CW}, Non-continuous)\end{tabular}} & \multicolumn{5}{c|}{\begin{tabular}[c]{@{}c@{}}Dataset \texttt{\ac{CI4R}}~\cite{DatasetB}\\ (\ac{FMCW}, Non-continuous)\end{tabular}} & \multicolumn{5}{c|}{\begin{tabular}[c]{@{}c@{}}Dataset \texttt{\ac{UoG20}}~\cite{DatasetC}\\ (\ac{FMCW}, Continuous)\end{tabular}} \\ \cline{2-16} 
    \multicolumn{1}{|c|}{\multirow{-3}{*}{Classifiers}} & \begin{tabular}[c]{@{}c@{}}\#params\\ (k)\end{tabular} & \begin{tabular}[c]{@{}c@{}}$\#$FLOP~\tnote{a} \\ (M)\end{tabular} & \begin{tabular}[c]{@{}c@{}}Accuracy\\ (\%)\end{tabular} & \begin{tabular}[c]{@{}c@{}}Peak Mem\\ (MB)\end{tabular} & \begin{tabular}[c]{@{}c@{}}Inf. Time\\ (ms)\end{tabular} & \begin{tabular}[c]{@{}c@{}}\#params\\ (k)\end{tabular} & \begin{tabular}[c]{@{}c@{}}$\#$FLOP.~\tnote{a} \\ (M)\end{tabular} & \begin{tabular}[c]{@{}c@{}}Accuracy\\ (\%)\end{tabular} & \begin{tabular}[c]{@{}c@{}}Peak Mem.\\ (MB)\end{tabular} & \begin{tabular}[c]{@{}c@{}}Inf. Time\\ (ms)\end{tabular} & \begin{tabular}[c]{@{}c@{}}\#params\\ (k)\end{tabular} & \begin{tabular}[c]{@{}c@{}}$\#$FLOP~\tnote{a} \\ (M)\end{tabular} & \begin{tabular}[c]{@{}c@{}}Accuracy\\ (\%)\end{tabular} & \begin{tabular}[c]{@{}c@{}}Peak Mem.\\ (MB)\end{tabular} & \begin{tabular}[c]{@{}c@{}}Inf. Time\\ (ms)\end{tabular} \\ \hline \hline
    ResNet6\tnote{b}~\cite{ResNet} & 69.4 & 253.3 & 99.8$\pm$0.16 & 11.9 & 4.8 & 150.7 & 310.2 & 32.3$\pm$3.35 & 12.8 & 5.1 & 24.7 & 72.8 & 83.6$\pm$1.73 & 10.7 & 4.8\\
    \ac{Bi}-\ac{LSTM}~\cite{DatasetC} & 23.3 & 5.6 & 93.9$\pm$0.68 & 10.5 & 1188.4 & 134.4 & 6.0 & 79.4$\pm$1.23 & 10.2 & 1173.4 & 15.5 & 2.1 & 85.3$\pm$2.01 & 9.7 & 1162.9 \\
    \ac{CNN}-\ac{LSTM}~\cite{CNN-LSTM} & 115.3 & 8.7 & 98.4$\pm$0.33 & 11.1 & 214.6 & 96.5 & 8.7 & 86.1$\pm$1.32 & 10.0 & 212.0 & 10.3 & 1.2 & 86.1$\pm$2.22 & 11.0 & 209.7 \\
    \ac{CNN}-\ac{Bi}-\ac{GRU}~\cite{CNN-RNN-distributed} & 55.0 & 1090.1 & 98.5$\pm$0.41 & 22.3 & 1314.5 & 142.8 & 1013.0 & 84.9$\pm$1.11 & 22.2 & 1305.5& 53.6 & 1082.0 & 85.4$\pm$2.19 & 22.8 & 1291.9 \\
    FML-ViT\tnote{c}~\cite{FML-VIT} & - & - & - & - & - & $\sim$2700 & 169.0 & $\sim$92.0 & - & - & - & - & - & - & -\\
    LW-VIT\tnote{c}~\cite{LW_VIT} & - & - & - & - & - & 769 & $\sim$2410 & $\sim$92.0 & - & - & - & - & - & - & - \\
    ActivityMamba\tnote{c}~\cite{ActivityMamba} & $\sim$8700.0 & $\sim$1220.0 & 99.8 & - & - & - & - & - & - & - & - & - & - & - & - \\ \hline \hline
    {\color[HTML]{9A0000} \textbf{RadMamba (Ours)}} & {\color[HTML]{9A0000} \textbf{21.7}} & {\color[HTML]{9A0000} \textbf{145.6}} & {\color[HTML]{9A0000} \textbf{99.8$\pm$0.12}} & {\color[HTML]{9A0000} \textbf{16.5}} & {\color[HTML]{9A0000}\textbf{21.2}} & {\color[HTML]{9A0000} \textbf{71.4}} & {\color[HTML]{9A0000} \textbf{8.8}} & {\color[HTML]{9A0000} \textbf{91.2$\pm$2.57}} & {\color[HTML]{9A0000} \textbf{10.2}} & {\color[HTML]{9A0000}\textbf{15.1}} & {\color[HTML]{9A0000} \textbf{6.7}} & {\color[HTML]{9A0000} \textbf{1.0}} & {\color[HTML]{9A0000} \textbf{89.3$\pm$1.38}} & {\color[HTML]{9A0000} \textbf{10.0}} & {\color[HTML]{9A0000}\textbf{11.6}} \\ {\color[HTML]{9A0000} \textbf{RadMamba (Ours)}} & {\color[HTML]{9A0000} \textbf{10.1}} & {\color[HTML]{9A0000} \textbf{14.5}} & {\color[HTML]{9A0000} \textbf{98.2$\pm$0.46}} & {\color[HTML]{9A0000} \textbf{16.5}} & {\color[HTML]{9A0000}\textbf{21.2}} & {\color[HTML]{9A0000} \textbf{47.9}} & {\color[HTML]{9A0000} \textbf{6.0}} & {\color[HTML]{9A0000} \textbf{88.8$\pm$4.58}} & {\color[HTML]{9A0000} \textbf{10.1}} & {\color[HTML]{9A0000}\textbf{14.9}} & {\color[HTML]{9A0000} \textbf{17.7}}  & {\color[HTML]{9A0000} \textbf{1.8 }}& {\color[HTML]{9A0000} \textbf{89.6$\pm$0.78}} & {\color[HTML]{9A0000}\textbf{10.0}} & {\color[HTML]{9A0000} \textbf{12.6}}\\\hline \hline
    \end{tabular}%
    \begin{tablenotes}
    \item[a] $\#$FLOP. is the number of floating-point operations per inference.
    \item[b] This ResNet6 only retains the first 5 layers and the output linear layers of ResNet18.
    \item[c] The '$\sim$' indicates that these results are approximated from the references~\cite{FML-VIT,LW_VIT,ActivityMamba}.
    \end{tablenotes}
    \end{threeparttable}}
    \end{table*}

\begin{figure}[!t]
    \centering
    \includegraphics[width=\linewidth]{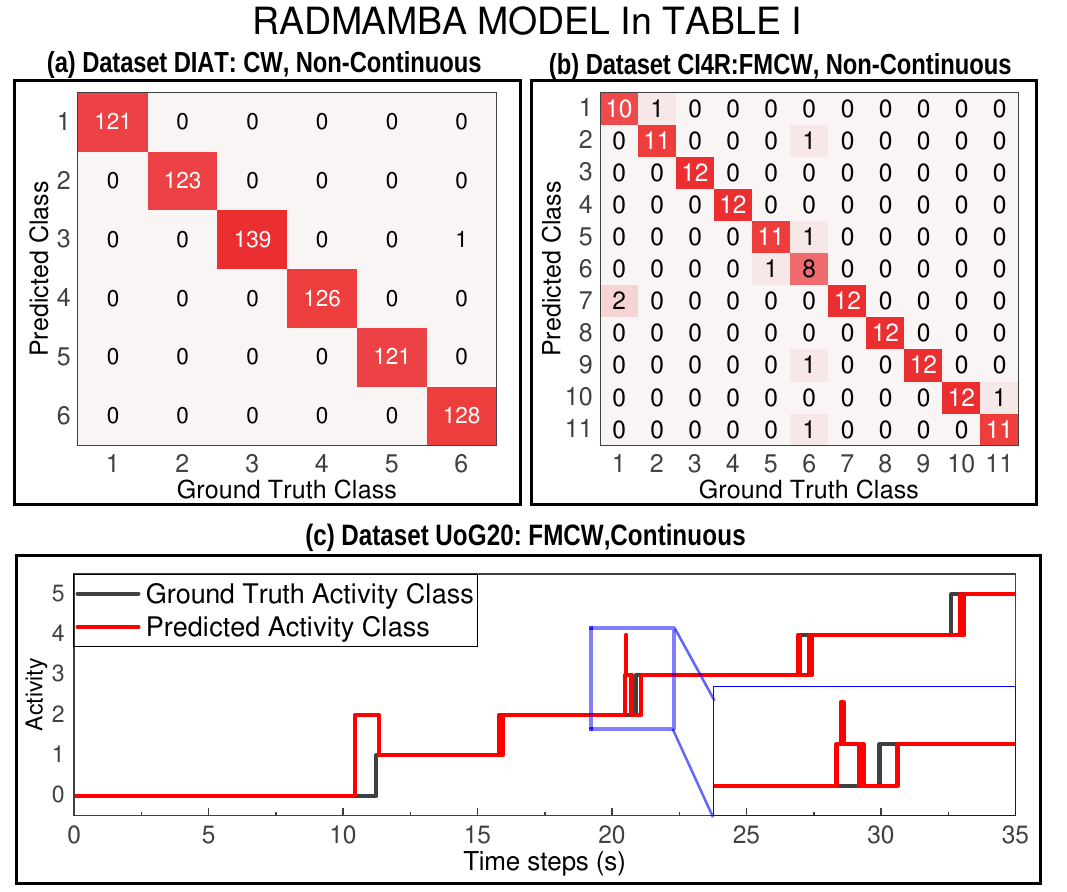}
    \caption{Example of comparisons between ground truth and best prediction of RadMamba model with the configuration in Table I on (a) Dataset \texttt{\ac{DIAT}}; (b) Dataset \texttt{\ac{CI4R}}; (c) Dataset \texttt{\ac{UoG20}}.}
    \label{fig:confusion}
\end{figure}

The experimental results detailed in Table~\ref{tab:model_comp} highlight the superior performance of our proposed RadMamba model across multiple radar-based human activity recognition datasets.

Figure~\ref{fig:confusion} (a) and (b) present confusion matrices comparing the ground truth labels with the best predictions of RadMamba, illustrating its classification accuracy and error distribution for each dataset. Figure~\ref{fig:confusion} (c) shows the comparison between the classification results of RadMamba and \ac{CNN}-\ac{LSTM} over time. 

On Dataset \texttt{\ac{DIAT}}, which utilizes CW radar in a non-continuous setting, RadMamba achieved an accuracy of 99.8\% with a standard deviation of $\pm$0.12\%, using only 21.7k parameters and requiring 145.6 million $\#$FLOP/Inf. per sample. This demonstrates its capability to match the state-of-the-art accuracy of previous models while being significantly more parameter-efficient compared to $\sim$8700k parameters and computationally lighter compared to $\sim$1220M $\#$FLOP per inference. The \ac{CNN}-\ac{LSTM} with 8.7\,M $\#$FLOP per inference achieves better accuracy of 98.4\% than 98.2\% from RadMamba with 14.5\,M $\#$FLOP per inference. 
Taking 45\,nm technology as an example, the energy consumption of an 8\,kB SRAM access for one 32-bit floating-point parameter and one 32-bit floating-point operation is 5.0\,pJ and 4.6\,pJ, respectively~\cite{horowitz20141}. With an 11$\times$ larger model and only about half the $\#$FLOPs per inference, it is unlikely that the \ac{CNN}-\ac{LSTM} can be more energy-efficient than RadMamba.

For Dataset \texttt{\ac{CI4R}}, employing \ac{FMCW} radar in a non-continuous scenario with 11 activities, RadMamba attained an accuracy of 91.2\% with a standard deviation of $\pm$2.57\%, leveraging 71.4 thousand parameters. This performance outperforms previous models shown in Tab.~\ref{tab:model_comp}. Moreover, it remains competitive with the state-of-the-art of previous models while requiring substantially fewer computational resources (8.8\,M $\#$FLOP/Inf. vs. 169\,M to 2410\,M $\#$FLOP/Inf., showcasing its ability to excel in more complex activity recognition tasks with minimal computational overhead. Moreover, compared to the \ac{Bi}-\ac{LSTM} model, which has the lowest arithmetic intensity among the baselines, RadMamba improves accuracy by 9.4 percentage points while using 86.5\,k fewer parameters and the same 6.0\,M \#FLOP/Inf.
%-------------------------------
\begin{figure*}[!t]
    \centering
    \includegraphics[width=\linewidth]{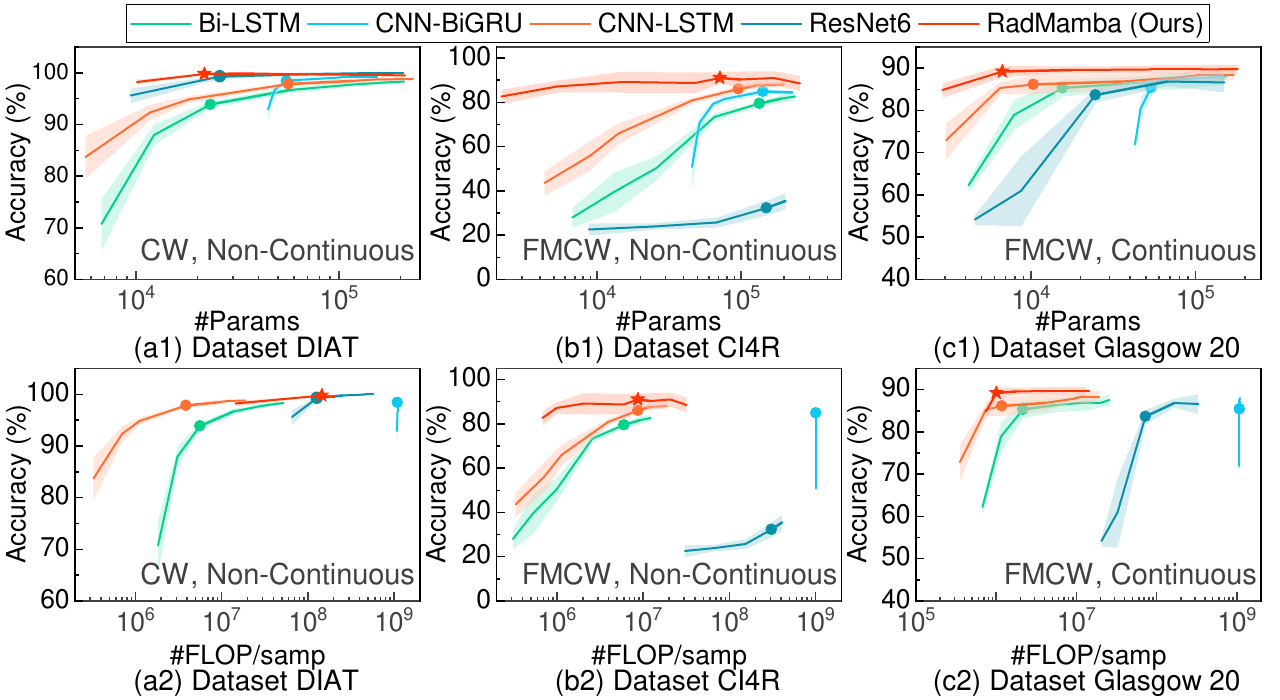}
    \caption{Accuracy vs. number of parameters (1) and number of $\#$FLOP/Inf. (2) on: (a) Dataset \texttt{\ac{DIAT}} (b) Dataset \texttt{\ac{CI4R}} (c) Dataset \texttt{\ac{UoG20}}}
    \label{fig:swept}
\end{figure*}
%-------------------------------

On Dataset \texttt{\ac{UoG20}}, which involves \ac{FMCW} radar in a continuous setting with 6 activities, RadMamba recorded an accuracy of 89.3\% with a standard deviation of $\pm$1.38\%, utilizing 6.7 thousand parameters and 1.0M $\#$FLOP/Inf. per inference. With an even smaller model size and a more complicated task than in Datasets \texttt{\ac{DIAT}} and \texttt{\ac{CI4R}}, RadMamba achieves the best accuracy among the models shown in Tab.~\ref{tab:model_comp} and highlights its effectiveness in continuous HAR scenarios with the smallest model size and $\#$FLOP/Inf. per inference.

Regarding inference time performance, as detailed in Tab.~\ref{tab:model_comp}, RadMamba demonstrates a substantial speedup compared to \ac{RNN}-based and hybrid models, achieving 10 to 100$\times$ faster inference. For instance, on Dataset \texttt{\ac{CI4R}}, RadMamba (71.4k parameters) achieves an inference time of 15.13\,ms. This represents a $78\times$ speedup over \ac{Bi}-\ac{LSTM} and an $87\times$ speedup over \ac{CNN}-\ac{Bi}-\ac{GRU}. While the conventional \ac{CNN} model, ResNet6, exhibits faster inference times (e.g., 5.14\,ms on the \texttt{\ac{CI4R}} dataset), its accuracy is significantly lower (32.3\% vs. RadMamba's 91.2\%). This result highlights RadMamba's superior trade-off between accuracy and efficiency, which is critical for reliable real-time \ac{RadHAR} applications.

We also measured peak memory consumption to address concerns about efficiency. The results in Tab.~\ref{tab:model_comp} reveal relatively small differences between models, with peak CUDA allocated memory ranging from 9.7\,MB to 22.8\,MB. This limited variation, despite RadMamba's significantly smaller parameter count, highlights a limitation of benchmarking on general-purpose, throughput-oriented embedded GPU architectures for edge inference when the batch size equals 1. On such platforms, the memory footprint is dominated by system-level overheads, such as CUDA context initialization, driver memory pools, and memory management, which mask the model's intrinsic parameter efficiency.

The true advantage of RadMamba's parameter efficiency is its suitability for future hardware-software co-design. Its minimal footprint, ranging from 6.7k to 71.4k parameters across the three tested datasets, is intended to enable the entire model to be buffered within on-chip SRAM on a dedicated \ac{RadHAR} accelerator. This approach would eliminate power-hungry off-chip DDR memory accesses, a primary source of power consumption in on-sensor systems. This fundamental power-saving benefit, along with the low arithmetic utilization for batch-size-1 inference on GPUs, cannot be captured by measurements on the Jetson platform. Therefore, we contend that the $\#$params and $\#$FLOPs metrics reported in Tab.~\ref{tab:model_comp} are more representative indicators of the RadMamba's potential for enabling ultra-low-power, on-sensor inference, which is the primary goal of this work.

\subsection{Model Scaling and Efficiency Analysis}
The model size swept results, illustrated in Figure~\ref{fig:swept}, compare accuracy against the number of parameters from 2k to 150k and $\#$FLOP/Inf. across Datasets \texttt{\ac{DIAT}}, \texttt{\ac{CI4R}}, and \texttt{\ac{UoG20}} between previous models ResNet6, \ac{Bi}-\ac{LSTM}, \ac{CNN}-\ac{Bi}-\ac{GRU}, and \ac{CNN}-\ac{LSTM} and the proposed RadMamba.  

For Dataset \texttt{\ac{DIAT}}, as presented in Fig.~\ref{fig:swept} (a1) and (a2), RadMamba achieves a peak accuracy of 99.8\% with only 21.7\,k parameters. However, RadMamba overfits with a model size larger than 21.7\,k because the dataset is very simple and the accuracy is close to 100\%. In contrast, ResNet6's accuracy increases steadily with model size, reaching 99.9\% with 208.1\,k parameters, yet it requires nearly ten times the parameters of RadMamba for slightly lower performance.

On Dataset \texttt{\ac{CI4R}}, as shown in Fig.~\ref{fig:swept} (b1) and (b2), RadMamba consistently outperforms previous models across a range of parameter configurations in efficiency, maintaining superior accuracy with fewer parameters and $\#$FLOP/Inf. Similarly, for Dataset \texttt{\ac{UoG20}} shown in Fig.~\ref{fig:swept} (c1) and (c2), RadMamba still records a higher accuracy for continuous activities scenarios with fewer $\#$FLOP/Inf., outperforming larger and more computationally intensive models in efficiency and effectiveness.

These trends, as depicted in the figure, affirm RadMamba's capability to achieve superior accuracy across diverse radar modalities while drastically reducing both parameter count and $\#$FLOP/Inf., positioning it as an optimal solution for lightweight, real-time human activity recognition. %to highlight the exceptional efficiency of our proposed RadMamba model by applications.

\subsection{Ablation study of the Micro-doppler-oriented Structure}
\begin{table*}[!t]
\caption{Ablation Study with progressive techniques (denoted by \textbf{Tech.}) of New Blocks in RadMamba across Seed 0 to 5, Evaluated with Three Different Radar Settings alongside Their Number of Parameters}
\label{tab:ablation}
\large{
\resizebox{\linewidth}{!}{%
\begin{threeparttable}
\begin{tabular}{|c|c|c|cccc|cccc|cccc|}
\hline \hline
 &  &  & \multicolumn{4}{c|}{Dataset \texttt{\ac{DIAT}}} & \multicolumn{4}{c|}{Dataset \texttt{\ac{CI4R}}} & \multicolumn{4}{c|}{Dataset \texttt{\ac{UoG20}}} \\ \cline{4-15} 
\multirow{-2}{*}{Row} & \multirow{-2}{*}{\begin{tabular}[c]{@{}c@{}} \textbf{Tech. 1}: Proj.\\ Layer\end{tabular}} & \multirow{-2}{*}{\begin{tabular}[c]{@{}c@{}} \textbf{Tech. 2}:\\ Patch Size\end{tabular}} & \begin{tabular}[c]{@{}c@{}}\textbf{Tech. 3}: \\ Down-sample.\end{tabular} & \multicolumn{2}{c}{\begin{tabular}[c]{@{}c@{}}Accuracy\\ (\%)\end{tabular}} & \begin{tabular}[c]{@{}c@{}}\#Params\\ (k)\end{tabular} & \begin{tabular}[c]{@{}c@{}}\textbf{Tech 3}: \\ Down-sample.\end{tabular} & \multicolumn{2}{c}{\begin{tabular}[c]{@{}c@{}}Accuracy\\ (\%)\end{tabular}} & \begin{tabular}[c]{@{}c@{}}\#Params\\ (k)\end{tabular} & \begin{tabular}[c]{@{}c@{}}\textbf{Tech 3}: \\ Down-sample.\end{tabular} & \multicolumn{2}{c}{\begin{tabular}[c]{@{}c@{}}Accuracy\\ (\%)\end{tabular}} & \begin{tabular}[c]{@{}c@{}}\#Param\\ (k)\end{tabular} \\ \hline \hline
1 (Conventional) &  &  & (1, 1) & 93.5 & $\pm$1.06 & 11.3 & (1, 1) & 61.9 & $\pm$2.55 & 40.6 & (1, 1) & 71.2 & $\pm$2.10 & 4.2 \\
2 &  &  & - & - & - & - & $\Updownarrow$ (8, 2) & 71.6 & $\pm$2.42 & 40.6 &  $\Updownarrow$ (8, 2) & 80.1 & $\pm$2.07 & 4.2 \\
3 &  & \multirow{-3}{*}{($H_{seg}$, $W_{seg}$)\tnote{a}} &  (2, 2) & 99.6 & $\pm$0.06 & 11.3 & $\Leftrightarrow$ (2, 8) & 63.0 & $\pm$7.75 & 40.6 &  $\Leftrightarrow$ (2, 32) & 66.9 & $\pm$2.01 & 4.2 \\ \cline{3-3}
4 &  &  & (1, 1) & 87.7 & $\pm$1.25 & 39.3 & (1, 1) & 67.9 & $\pm$0.76 & 54.9 & (1, 1) & 72.6 & $\pm$2.20 & 7.6 \\
5 &  &  & - & - & - & - & $\Updownarrow$ (8, 2) & 65.2 & $\pm$2.73 & 45.8 &  $\Updownarrow$ (8, 2) & 67.7 & $\pm$1.27 & 5.5 \\
6 &  & \multirow{-3}{*}{(1, $W_{c-d}$)} &  (2,  2) & 98.5 & $\pm$0.43 & 21.4 & $\Leftrightarrow$ (2, 8) & 77.3 & $\pm$2.80 & 38.9 &  $\Leftrightarrow$ (2, 32) & 64.0 & $\pm$1.26 & 3.6 \\ \cline{3-3}
7 &  &  & (1, 1) & 96.9 & $\pm$0.52 & 39.3 & (1, 1) & 88.4 & $\pm$0.76 & 54.9 & (1, 1) & 83.9 & $\pm$1.85 & 7.8 \\
8 &  &  & - & - & - & - & $\Updownarrow$ (8, 2) & 84.6 & $\pm$4.12 & 38.9 &  $\Updownarrow$ (8, 2) & 84.3 & $\pm$1.88 & 4.1 \\
9 & \multirow{-9}{*}{\begin{tabular}[c]{@{}c@{}}1 linear\\ layer\end{tabular}} & \multirow{-3}{*}{($H_{c-d}$, 1)} &  (2, 2) & 99.9 & $\pm$0.00 & 21.4 & $\Leftrightarrow$ (2, 8) & 86.5 & $\pm$4.35 & 45.8 &  $\Leftrightarrow$ (2, 32) & 84.2 & $\pm$2.00 & 5.7 \\ \hline
10 &  &  & (1, 1) & 92.6 & $\pm$0.98 & 11.6 & (1, 1) & 40.1 & 11.98 & 66.2 & (1, 1) & 70.6 & $\pm$2.12 & 5.2 \\
11 &  &  & - & - & - & - & $\Updownarrow$ (8, 2) & 69.9 & $\pm$2.99 & 66.2 &  $\Updownarrow$ (8, 2) & 78.2 & $\pm$3.66 & 5.2 \\
12 &  & \multirow{-3}{*}{($H_{seg}$, $W_{seg}$)\tnote{a}} &  (2, 2) & 99.5 & $\pm$0.10 & 11.6 & $\Leftrightarrow$ (2, 8) & 57.8 & $\pm$9.64 & 66.2 &  $\Leftrightarrow$ (2, 32) & 68.5 & $\pm$1.50 & 5.2 \\ \cline{3-3}
13 &  &  & (1, 1) & 88.8 & $\pm$2.42 & 39.6 & (1, 1) & 48.3 & $\pm$9.80 & 80.5 & (1, 1) & 70.8 & $\pm$2.43 & 8.6 \\
14 &  &  & - & - & - & - & $\Updownarrow$ (8, 2) & 43.3 & $\pm$8.54 & 71.4 &  $\Updownarrow$ (8, 2) & 65.2 & $\pm$1.48 & 6.6 \\
15 &  & \multirow{-3}{*}{(1, $W_{c-d}$)} &  (2, 2) & 98.7 & $\pm$0.32 & 21.7 & $\Leftrightarrow$ (2, 8) & 72.4 & $\pm$3.12 & 64.5 &  $\Leftrightarrow$ (2, 32) & 64.0 & $\pm$2.16 & 4.7 \\ \cline{3-3}
16 &  &  & (1, 1) & 97.2 & $\pm$0.26 & 39.6 & (1, 1) & 85.3 & $\pm$1.27 & 80.5 & (1, 1) & 83.6 & $\pm$1.04 & 8.9 \\
17 &  &  & - & - & - & - & $\Updownarrow$ (8, 2) & 83.9 & $\pm$1.71 & 64.5 &  $\Updownarrow$ (8, 2) & 83.2 & $\pm$1.64 & 5.1 \\
18 & \multirow{-9}{*}{\begin{tabular}[c]{@{}c@{}}3 linear\\ layer\tnote{b}\end{tabular}} & \multirow{-3}{*}{($H_{c-d}$, 1)} &  (2, 2) & 99.8 & $\pm$0.06 & 21.7 & $\Leftrightarrow$ (2, 8) & 85.3 & $\pm$2.67 & 71.4 &  $\Leftrightarrow$ (2, 32) & 85.2 & $\pm$1.16 & 6.7 \\ \hline
19 &  &  & (1, 1) & 93.8 & $\pm$1.14 & 11.6 & (1, 1) & 65.5 & $\pm$3.30 & 66.2 & (1, 1) & 75.3 & $\pm$2.34 & 5.2 \\
20 &  &  & - & - & - & - & $\Updownarrow$ (8, 2) & 75.0 & $\pm$2.45 & 66.2 &  $\Updownarrow$ (8, 2) & 81.0 & $\pm$2.10 & 5.2 \\
21 &  & \multirow{-3}{*}{($H_{seg}$, $W_{seg}$)\tnote{a}} &  (2, 2) & 99.6 & $\pm$0.15 & 11.6 & $\Leftrightarrow$ (2, 8) & 70.7 & $\pm$3.23 & 66.2 &  $\Leftrightarrow$ (2, 32) & 74.3 & $\pm$2.31 & 5.2 \\ \cline{3-3}
22 &  &  & (1, 1) & 90.1 & $\pm$1.72 & 39.6 & (1, 1) & 70.7 & $\pm$3.43 & 80.5 & (1, 1) & 76.4 & $\pm$2.77 & 8.6 \\
23 &  &  & - & - & - & - & $\Updownarrow$ (8, 2) & 69.3 & $\pm$3.09 & 71.4 &  $\Updownarrow$ (8, 2) & 74.4 & $\pm$3.26 & 6.6 \\
24 &  & \multirow{-3}{*}{(1, $W_{c-d}$)} &  (2, 2) & 98.9 & $\pm$0.13 & 21.7 & $\Leftrightarrow$ (2, 8) & 79.0 & $\pm$2.42 & 64.5 &  $\Leftrightarrow$ (2, 32) & 68.8 & $\pm$2.08 & 4.7 \\ \cline{3-3}
25 &  &  & (1, 1) & 96.8 & $\pm$0.86 & 39.6 & (1, 1) & 87.9 & $\pm$0.36 & 80.5 & (1, 1) & 84.5 & $\pm$0.77 & 8.9 \\
26 &  &  & - & - & - & - &  $\Updownarrow$ (8, 2) &  85.6 &  $\pm$3.16 &  64.5 &  $\Updownarrow$ (8, 2)& 84.4 & $\pm$2.05 & 5.1 \\
{\color[HTML]{9A0000} 27 (Ours)} & \multirow{-9}{*}{\begin{tabular}[c]{@{}c@{}}1 conv1d\\ layer\tnote{c}\end{tabular}} & \multirow{-3}{*}{($H_{c-d}$, 1)} & {\color[HTML]{9A0000} (2, 2)} & {\color[HTML]{9A0000} 99.8} & {\color[HTML]{9A0000} $\pm$0.11} & {\color[HTML]{9A0000} 21.7} & {\color[HTML]{9A0000} $\Leftrightarrow$ (2, 8)} & {\color[HTML]{9A0000} 89.6} & {\color[HTML]{9A0000} $\pm$3.69} & {\color[HTML]{9A0000} 71.4} & {\color[HTML]{9A0000} $\Leftrightarrow$ (2, 32)} & {\color[HTML]{9A0000} 88.6} & {\color[HTML]{9A0000} $\pm$1.46} & {\color[HTML]{9A0000} 6.7} \\ \hline \hline
\end{tabular}%
\begin{tablenotes}
\large{
\item[a] $(H_1, W_1)$ equals to (7, 7) for Dataset \texttt{\ac{DIAT}} and B, (5, 7) for Dataset \texttt{\ac{UoG20}} due to the different image size of these datasets.
\item[b] This option replaces projections 1 and 2 from a single linear layer to three linear layers. 
\item[c] The kernel size for projection 1 and 2 is $(dim, dim, 3)$, and for projection 3 is $(dim, dim, 1)$.}
\end{tablenotes}
\end{threeparttable}}}
\end{table*}

Table~\ref{tab:ablation} presents an ablation study evaluating the impact of new architectural blocks in the RadMamba model across three datasets with varying radar settings. The study systematically assesses the effects of different projection layers, patch sizes, and downsampling scales on model accuracy and parameter efficiency, with results averaged over seeds 0 to 5. 

The main conclusion from this ablation study is that optimal performance emerges from the combined use of all three proposed new blocks, including the projection layer, the patch size, and the downsampling scale, rather than any single component in isolation. This is evident in the incremental improvements observed across the configurations. For instance, the baseline conventional \ac{ViM} model (Row 1) achieves an accuracy of 93.5\% on Dataset \texttt{\ac{DIAT}}, 61.9\% on Dataset \texttt{\ac{CI4R}}, and 71.2\% on Dataset \texttt{\ac{UoG20}}. In contrast, the proposed RadMamba (Row 27), which incorporates a 1D convolution layer, a patch size of ($H_{c-d}$, 1), and a downsampling (2, 2), achieves a higher accuracy of 99.8\% on Dataset \texttt{\ac{DIAT}}. Similarly, on Dataset \texttt{\ac{CI4R}}, Row 27 with a (2, 8) downsampling scale yields significantly higher accuracy of 89.6\%, and on Dataset \texttt{\ac{UoG20}} with a (2, 32) scale, it reaches 88.6\%. Intermediate configurations, such as Row 19 (only projection replacement), Row 7 (only patch size changing), and Row 2 (only downsampling applied) show lower accuracies—underscoring that no single structure alone achieves the peak performance of the fully combined design in Row 27.

To provide a more detailed analysis, we discuss the three blocks—projection layer, patch size, and downsampling scale—separately, evaluating their individual contributions to RadMamba's performance across the datasets.

\subsubsection{Projection Replacement}
The adoption of a 1D convolutional layer as the projection mechanism consistently outperforms configurations with a single linear layer or three linear layers across nearly all settings. Comparing Rows 19-27 with Rows 10-18 and Rows 1-9, the conv1d projection replacement within the same patch size and downsampling scale achieved an average increment of mean accuracy by 0.48\% over 1 linear layer, 0.40\% over 3 linear layers on Dataset \texttt{\ac{DIAT}}.
On datasets \texttt{\ac{CI4R}} and \texttt{\ac{UoG20}}, the conv1d layer shows a substantial improvement of 2.99\% and 3.64\% over 1 linear layer, respectively.

\subsubsection{Patch Size}

The ($H_{c-d}$, 1) configuration consistently delivers superior performance within each projection layer type. For the 1D convolutional layer, Row 25 with ($H_{c-d}$, 1) and without downsampling achieves 3\% gain on Dataset \texttt{\ac{DIAT}}, 22.4\% gain on Dataset \texttt{\ac{CI4R}}, 9.2\% gain on Dataset \texttt{\ac{UoG20}} vs. Row 19 with rectangular patch. This trend holds across other projection types, indicating that the ($H_{c-d}$, 1) patch size effectively retains the critical spatial features for radar data processing.

\subsubsection{Downsampling}
The results suggest that downsampling along the patch vector dimension generally outperforms downsampling along the dimension orthogonal to the patch vector and no downsampling. For instance, Row 27's (2, 32) time downsampling achieves 88.6\% $\pm$ 1.46, outperforming Row 26's (8, 2) at 84.4\% $\pm$ 2.05 and Row 25's (1, 1) at 84.5\% $\pm$ 0.77 on Dataset \texttt{\ac{UoG20}}. This pattern highlights the advantage of downsampling in enhancing the micro-Doppler features.

The ablation study underscores the efficacy of the proposed RadMamba design, with the highest performance achieved by integrating a 1D convolutional projection layer, a ($H_{c-d}$, 1) patch size, and time-dimension downsampling (Row 27).

\section{Conclusion}
RadMamba sets a new benchmark in RadHAR ({\color{red}Code: \url{https://github.com/lab-emi/AIRHAR}}), delivering top-tier accuracy with minimal parameters and computational cost across diverse datasets. Its radar micro-Doppler-oriented design addresses domain-specific challenges, making it ideal for real-time, resource-constrained applications.
Overall, RadMamba consistently outperforms previous neural network models like ResNet6, \ac{Bi}-\ac{LSTM}, \ac{CNN}-\ac{LSTM}, and \ac{CNN}-\ac{Bi}-\ac{GRU}, and competes with or exceeds advanced architectures like Vision Transformers and ActivityMamba, all while maintaining a remarkably low parameter count (ranging from 6.7\,k to 71.4\,k) and reduced computational complexity (1.3\,M to 145.6\,M $\#$FLOP/Inf.). These results emphasize RadMamba's capability as a lightweight and high-performance solution for radar-based HAR across diverse radar modalities and activity recognition tasks. 
The micro-Doppler-oriented design, which integrates channel fusion, Doppler-aligned segmentation, and convolutional projections, addresses the unique challenges of radar data, delivering to deliver high accuracy across both \ac{CW} and \ac{FMCW} datasets, both including continuous and non-continuous data. Ablation studies confirm the synergy of its components, with no single element achieving comparable performance on its own. 

\section*{Acknowledgments}
This work is funded by the European Research Executive Agency (REA) under the Marie Skłodowska-Curie Actions (MSCA) Postdoctoral Fellowship program, Grant No. 101107534 (AIRHAR). We thank Prof. Dr. L.C.N. de Vreede and Prof. Dr. Alexander Yarovoy for their support.

\bibliographystyle{IEEEtran}
\bibliography{IEEEabrv,reference} 

\vspace{-30pt}
\begin{IEEEbiography}[{\includegraphics[width=1in,height=1.25in,clip,keepaspectratio]{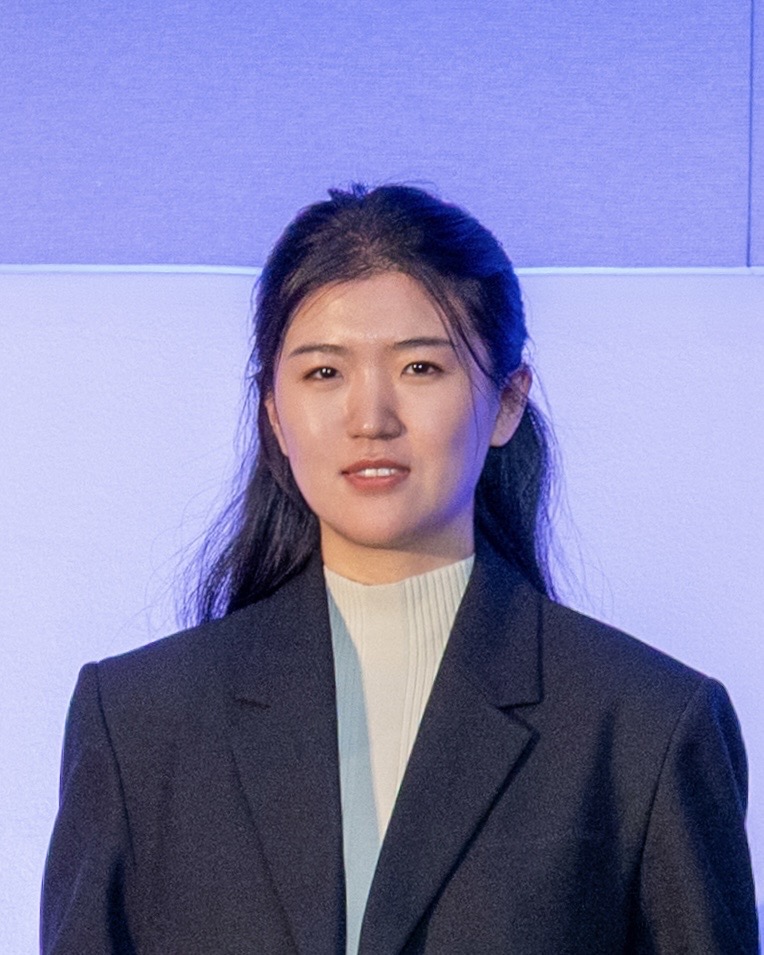}}]{Yizhuo Wu} (Student Member, IEEE) obtained her M.Sc. degree in Microelectronics at TU Delft in 2023. She is now a PhD student supervised by Dr. Chang Gao in the Lab of Efficient Machine Intelligence (EMI). Her research focuses on software-hardware co-designed AI for I/Q signal processing, which aims to find energy-efficient solutions for high-frequency signal processing tasks.
\end{IEEEbiography}
\vspace{-35pt}
\begin{IEEEbiography}[{\includegraphics[width=1in,height=1.25in,clip,keepaspectratio]{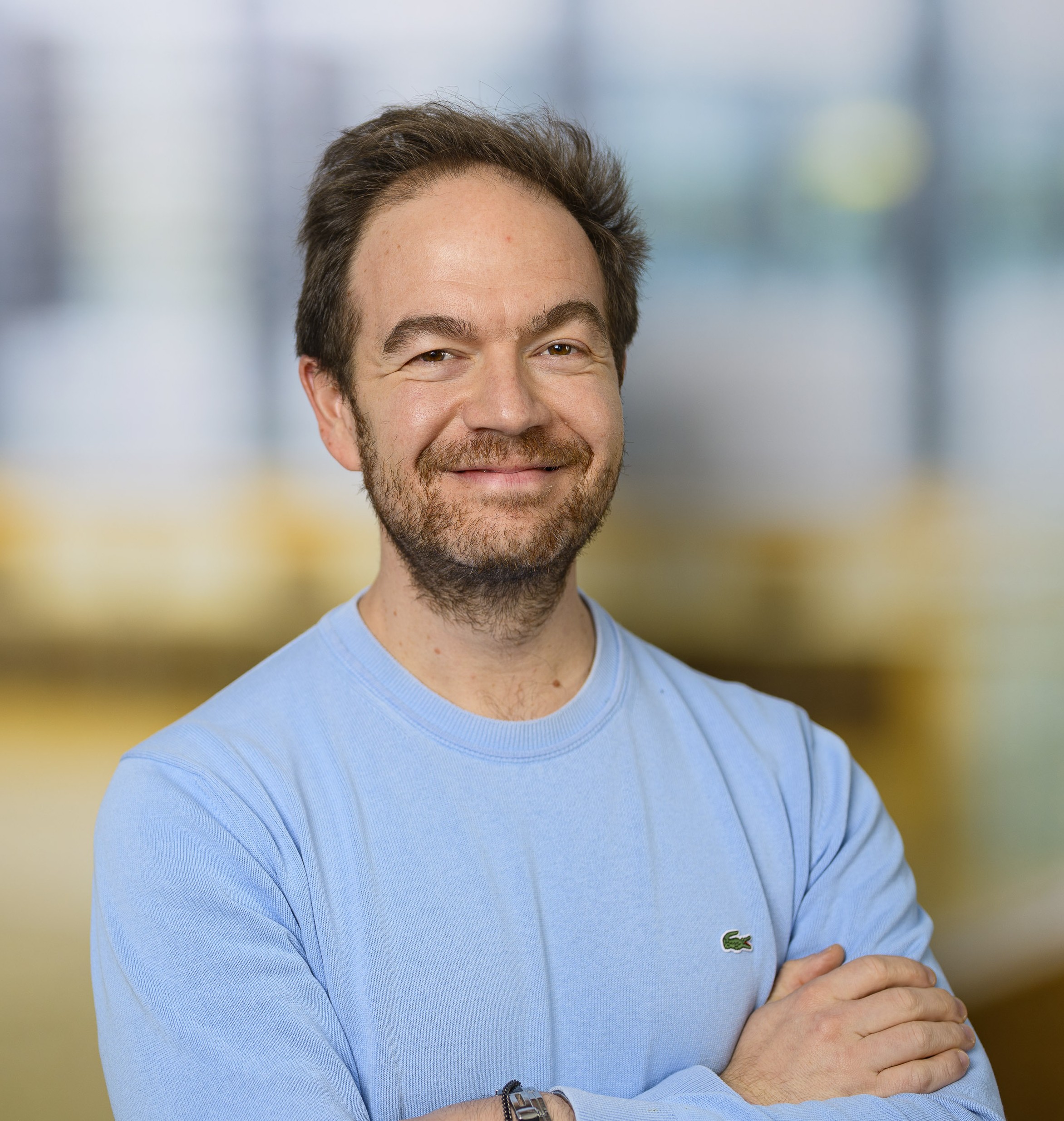}}]{Francesco Fioranelli}
(Senior Member, IEEE)
received the Laurea (B.Eng., cum laude) and Laurea Specialistica (M.Eng., cum laude) degrees in
telecommunication engineering from the Università Politecnica delle Marche, Ancona, Italy, in
2007 and 2010, respectively, and the Ph.D. degree
from Durham University, U.K., in 2014.
He is currently an Associate Professor with TU
Delft, Delft, The Netherlands. He was a Research
Associate at University College London, London,
U.K., from 2014 to 2016, and an Assistant Professor
at the University of Glasgow, Glasgow, U.K., from 2016 to 2019. He has
authored over 190 peer-reviewed publications and edited the books titled
Micro-Doppler Radar and Its Applications and Radar Countermeasures for
Unmanned Aerial Vehicles published by IET-Scitech in 2020. His research
interests include the development of radar systems and automatic classification
for human signatures analysis in healthcare and security, drone and UAV
detection and classification, automotive radar, wind farms, and sea clutter.
Dr. Fioranelli received four best paper awards and the IEEE AESS Fred
Nathanson Memorial Radar Award in 2024.
\end{IEEEbiography}
\vspace{-35pt}
\begin{IEEEbiography}[{\includegraphics[width=1in,height=1.25in,clip,keepaspectratio]{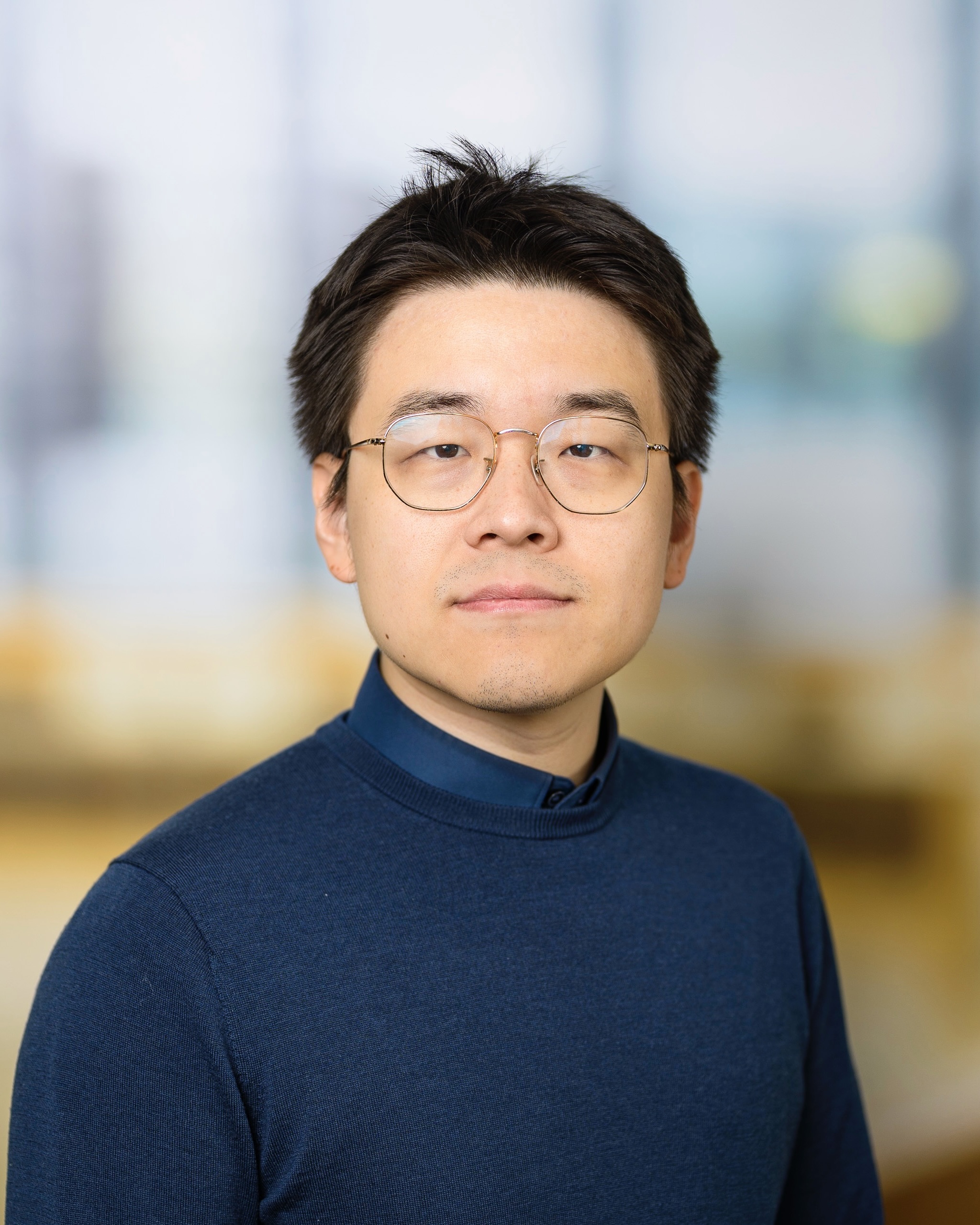}}]{Chang Gao} (Member, IEEE) received his Ph.D. degree with distinction in Neuroscience from the Institute of Neuroinformatics, University of Zürich and ETH Zürich, Zürich, Switzerland, in March 2022. He received his M.Sc. degree from Imperial College London in September 2016 and his B.Eng. degree from the University of Liverpool and Xi’an Jiaotong–Liverpool University in July 2015. In August 2022, he joined Delft University of Technology, The Netherlands, as a tenured Assistant Professor in the Department of Microelectronics. He leads the Lab of Efficient Machine Intelligence (EMI), where he conducts research on hardware–software co-design for edge AI computing and embodied intelligence. He received the 2022 Misha Mahowald Early Career Award in Neuromorphic Engineering and a 2022 Marie Skłodowska-Curie Postdoctoral Fellowship. He is a 2023 Dutch Research Council (NWO) Veni laureate and a 2023 MIT Technology Review Innovator Under 35 in Europe for his contributions to algorithm–hardware co-design for efficient sparse recurrent neural-network edge computing.
\end{IEEEbiography}

\end{document}